\documentclass[conference,usletter]{IEEEtran}
\usepackage{times,amsmath,amssymb}
\usepackage{slashbox}
\usepackage{graphicx}
\usepackage{xcolor}
\usepackage{algorithm}
\usepackage{algorithmicx}
\usepackage{setspace}
\usepackage[export]{adjustbox}
\usepackage{multirow}
\usepackage[noend]{algpseudocode}
\usepackage{tabularx,booktabs}
\usepackage{tikz}
\usepackage{makecell}
\DeclareMathOperator*{\argmin}{arg\,min}
\begin{document}

\title{Multilinear Compressive Learning with Prior Knowledge}
\author{\IEEEauthorblockN{Dat Thanh Tran\IEEEauthorrefmark{1}, Moncef Gabbouj\IEEEauthorrefmark{1}, Alexandros Iosifidis\IEEEauthorrefmark{2}}
\IEEEauthorblockA{\IEEEauthorrefmark{1}Department of Computing Sciences, Tampere University, Tampere, Finland\\
\IEEEauthorrefmark{2}Department of Engineering, Aarhus University, Aarhus, Denmark\\
Email:\{thanh.tran, moncef.gabbouj\}@tuni.fi, alexandros.iosifidis@eng.au.dk}\\

}

\maketitle

\begin{abstract}
The recently proposed Multilinear Compressive Learning (MCL) framework combines Multilinear Compressive Sensing and Machine Learning into an end-to-end system that takes into account the multidimensional structure of the signals when designing the sensing and feature synthesis components. The key idea behind MCL is the assumption of the existence of a tensor subspace which can capture the essential features from the signal for the downstream learning task. Thus, the ability to find such a discriminative tensor subspace and optimize the system to project the signals onto that data manifold plays an important role in Multilinear Compressive Learning. In this paper, we propose a novel solution to address both of the aforementioned requirements, i.e., \textit{How to find those tensor subspaces in which the signals of interest are highly separable?} and \textit{How to optimize the sensing and feature synthesis components to transform the original signals to the data manifold found in the first question?} In our proposal, the discovery of a high-quality data manifold is conducted by training a nonlinear compressive learning system on the inference task. Its knowledge of the data manifold of interest is then progressively transferred to the MCL components via multi-stage supervised training with the supervisory information encoding how the compressed measurements, the synthesized features, and the predictions should be like. The proposed knowledge transfer algorithm also comes with a semi-supervised adaption that enables compressive learning models to utilize unlabeled data effectively. Extensive experiments demonstrate that the proposed knowledge transfer method can effectively train MCL models to compressively sense and synthesize better features for the learning tasks with improved performances, especially when the complexity of the learning task increases. 
\end{abstract}

\section{Introduction}\label{S1}

Compressive Sensing (CS) \cite{candes2008introduction} is an efficient signal acquisition paradigm that performs the measurement of the signal at sub-Nyquist rates by sensing and linearly interpolating the samples at the sensor level. Due to the property that the compression step is performed before signal registration during the sampling phase, CS significantly lowers the temporary storage and bandwidth requirement of the sensing devices. Therefore, this paradigm plays an important role in many applications that involve high-dimensional signals since the signal collection process can be very computationally demanding, time-consuming, or even prohibitive when using the traditional point-based sensing-then-compressing approach \cite{caiafa2013multidimensional}. For example, Hyperspectral Compressive Imaging (HCI) \cite{august2013compressive} and Synthetic Aperture Radar (SAR) imaging \cite{patel2010compressed} often perform remote sensing of large volume of data constrained by limited electrical power, storage capacity and transmission bandwidth, or in 3D Magnetic Resonance Imaging (MRI) \cite{irarrazabal1995fast}, long scanning time can make patients feel uncomfortable, promoting body movements, thus degrading the image quality. 

Although the ideal sampling theorem requires the signal to be sampled at higher rates than the Nyquist rate to ensure perfect reconstruction, in CS, the undersampled signal can still be reconstructed almost perfectly if the sparsity assumption holds and the sensing operators possess certain properties \cite{candes2006stable,donoho2006compressed}. The sparsity prior assumes the existence of a set of basis or a dictionary in which the signal of interest has sparse representations, i.e., having few non-zero coefficients. Sparsity manifests in many classes of signals that we operate on since they are often smooth or piece-wise smooth, thus having sparse representations in Fourier or wavelet domains. This form of prior knowledge incorporated into the model allows us to acquire the signals at a much lower cost. 

In fact, over the past decade, a large body of works in CS has been dedicated to \textit{finding} relevant priors from the class of signals of interest, and \textit{incorporating} them into the model to achieve better compression and reconstruction \cite{mota2017compressed}. A typical form of priors that has been widely used in CS is the existence of a signal similar to the original signal of interest. This information is available in many scenarios such as video acquisitions \cite{stankovic2009compressive,kang2009distributed} and estimation problems \cite{charles2011sparsity} where signals acquired in succession are very similar. Similarity also exists in signals captured by nearby sensors in sensor networks \cite{baron2005distributed} or images from spatially close cameras in multiview camera systems \cite{cevher2008compressive,trocan2010disparity}. Other examples of prior information include the available estimate of the support of the signal \cite{vaswani2010modified}, Gaussian mixture models \cite{renna2014reconstruction}, weighted sparsity \cite{khajehnejad2011analyzing} or block sparsity \cite{stojnic2009reconstruction,eldar2009robust}. The process of \textit{finding} prior information often relies on the available prior knowledge related to the signal, e.g., the spatial arrangement of sensor networks, while the \textit{incorporation} of such information is often expressed as another term to be minimized such as the difference between the current frame and the last frame when reconstructing MRI images. For a detailed discussion on prior information in CS and its theoretical and empirical improvements, we refer the readers to \cite{mota2017compressed}. 

While signal reconstruction is a necessary processing step in some applications, there exist many scenarios in which the primary purpose is to detect certain patterns or to infer some properties from the acquired signal, rather than the recovery of the entire signal. Learning from compressed measurements, also known as Compressive Learning (CL) \cite{calderbank2012finding,davenport2007smashed,davenport2010signal,reboredo2013compressive,lohit2015reconstruction,adler2016compressed,tran2019multilinear}, focuses on the task of inference directly from compressive domain without explicit signal reconstruction. That is, CL models are often formed under the assumption that only certain aspects or representative features of the signal need to be captured to make inferences, even when perfect signal recovery might be impossible from the retained information. It is worth noting that in many scenarios such as security and surveillance applications \cite{yamac2019reversible,mohassel2017secureml}, signal recovery might disclose privacy information. Thus, it is undesirable. 

Since the main objective of CL is to extract relevant information given the learning problem, the literature in CL mainly concerns with the selection of the sensing operators \cite{reboredo2013projections,baheti2008adaptive} or the optimization of the learning model \cite{davenport2007smashed} that operates in the compressive domain. In the early works, the selection or design of sensing matrices was decoupled from the optimization of the inference model. Nowadays, with the developments on both hardware and algorithmic levels, stochastic optimization has become more and more efficient, making end-to-end learning systems applicable and appealing to a wide range of applications. Initiated by the work of \cite{lohit2016direct,adler2016compressed}, recent CL systems \cite{hollis2018compressed,deugerli2018compressively,zisselman2018compressed,tran2019multilinear} have adopted an end-to-end approach which jointly optimizes the sensing matrices and the inference operators, leaving the task of discovering relevant features to stochastic gradient descent. 

The majority of CL systems employs linear sensing operators which operate on the vector-based input, regardless of the natural representation of the signal. Recently, the authors in \cite{tran2019multilinear} proposed Multilinear Compressive Learning (MCL) framework, which takes into consideration the natural structure of multidimensional signals by adopting multilinear sensing and feature extraction components. MCL has been shown both theoretically and empirically to be more efficient than its vector-based counterpart. The assumption made in MCL is that the original signals can be linearly projected along each tensor mode to a tensor subspace which retains relevant information for the learning problem, and from which discriminative features can be synthesized. In a general setting, the authors in \cite{tran2019multilinear} propose to initialize the sensing matrices with the left singular vectors obtained from Higher Order Singular Value Decomposition (HOSVD), a kind of \textit{weak prior} that initially guides the model towards an energy-preserving tensor subspace during stochastic optimization.

As we have seen from CS literature, finding good priors and successfully incorporating them into the signal model lead us to better solutions. While prior information exists in many forms in the reconstruction task, it is not straightforward to define relevant information given the learning problem. Indeed, representation learning \cite{bengio2013representation} has been an active research area that aims to extract meaningful representations from the raw data, ideally without any human given label such as object categories. From the human perception, in certain cases, we do have an idea of what kind of information is relevant or irrelevant to the learning task, e.g., color cues might be irrelevant in face recognition problems but relevant for the traffic light detection problem. From the optimization point of view, it is hard to make such a claim before any trial. Therefore, we often want to strike a balance between hand-crafted features and end-to-end solutions. 

In this paper, we aim to tackle the problems of \textit{How to find good priors in Compressive Learning?} and \textit{How to incorporate such priors into an end-to-end Compressive Learning system without hard-coding?} Based on two observations:

\begin{itemize}
\item Although the sensing operators in CL are limited to linear/multilinear form, the feature synthesis component that operates on compressed measurements can adopt nonlinear transformations.
\item Nonlinear sensing operators have better capacities to discover representative tensor subspaces   
\end{itemize} 
we propose a novel approach that utilizes nonlinear compressive learning models to discover the structures of the compressive domain and the informative features that should be synthesized from this domain for the learning task. This knowledge is subsequently transferred to the compressive learning models via progressive Knowledge Distillation \cite{hinton2015distilling}. Our contributions can be summarized as follows:

\begin{itemize}
\item In this paper, a novel methodology to discover and incorporate prior knowledge into existing end-to-end Compressive Learning systems is proposed. Although we limit our investigation of the proposed method to Multilinear Compressive Learning framework, the approach described in this paper can be applied to any end-to-end Compressive Learning system to improve its performance. 

\item The proposed approach naturally leads to the semi-supervised extension in which the availability of unlabeled signals coming from similar distributions can benefit Compressive Learning systems via prior knowledge incorporation. 

\item We carefully designed and conducted extensive experiments in object classification and face recognition tasks to investigate the effectiveness of the proposed approach to learn and incorporate prior information. In addition, to facilitate future research and reproducibility, we publicly provide our implementation of all experiments reported in this paper\footnote{https://github.com/viebboy/MultilinearCompressiveLearningWithPrior}. 
\end{itemize}

The remainder of our paper is organized as follows: in Section \ref{S2}, we review the related works in Compressive Learning, Knowledge Distillation, and give a brief description of Multilinear Compressive Learning framework. Section \ref{S3} provides a detailed description of our approach to learn and incorporate prior knowledge into MCL models in both supervised and semi-supervised settings. Section \ref{S4} details our empirical analysis, including experiment protocols, experiment designs, and quantitative and qualitative evaluation. Section \ref{S5} concludes our work.

\section*{Abbreviation \& Nomenclature}
\addcontentsline{toc}{section}{Abbreviation \& Nomenclature}
\begin{IEEEdescription}[\IEEEusemathlabelsep\IEEEsetlabelwidth{$V_1,V_2,V_3$}]
	\item[CS] Compressive Sensing
	\item[CL] Compressive Learning
	\item[MCL] Multilinear Compressive Learning
	\item[MCLwP] Multilinear Compressive Learning with Priors
	\item[HOSVD] Higher Order Singular Value Decomposition
	\item[KD] Knowledge Distillation
	\item[FS] Feature Synthesis 
	\item[$\mathcal{Y}$] Multi-dimensional analog signal
	\item[$\mathbf{z}$] Compressed measurements (vector) in CL
	\item[$\mathcal{Z}$] Compressed measurements (tensor) in MCL
	\item[$\mathcal{T}$] Tensor features, output of FS component in MCL
	\item[$\mathsf{E}$] CS component in MCL
	\item[$\theta_{\mathsf{E}}$] Parameters of $\mathsf{E}$
	\item[$\mathsf{D}$] FS component in MCL
	\item[$\theta_{\mathsf{D}}$] Parameters of $\mathsf{D}$
	\item[$\mathsf{N}$] Task-specific neural network in MCL
	\item[$\theta_{\mathsf{N}}$] Parameters of $\mathsf{N}$
	\item[$\mathsf{P}$] Prior-generating model
	\item[$\mathcal{Z}_{\mathsf{P}}$] Compressed measurements in $\mathsf{P}$
	\item[$\mathcal{T}_{\mathsf{P}}$] Tensor features, output of FS component in $\mathsf{P}$
	\item[$\mathsf{E_P}$] Nonlinear sensing component in $\mathsf{P}$
	\item[$\theta_{\mathsf{E_P}}$] Parameters of $\mathsf{E_P}$
	\item[$\mathsf{D_P}$] FS component in $\mathsf{P}$
	\item[$\theta_{\mathsf{D_P}}$] Parameters of $\mathsf{D_P}$
	\item[$\mathsf{N_P}$] Task-specific neural network in $\mathsf{P}$
	\item[$\theta_{\mathsf{N_P}}$] Parameters of $\mathsf{N_P}$
	\item[$\mathcal{Y}_i$] $i$-th data sample
	\item[$c_i$] label of the $i$-th sample
	\item[$N$] Number of labeled samples
	\item[$M$] Total number of samples, including labeled and unlabeled samples
	\item[$\mathfrak{L}$] The set of labeled data
	\item[$\mathfrak{U}$] The set of unlabeled data
	\item[$\tilde{\mathfrak{L}}$] Enlarged labeled set
	\item[$L_I$] Inference loss function
	\item[$L_D$] Distillation loss function
	\ 
\end{IEEEdescription}

\section{Related Work}\label{S2}

\subsection{Background}

Throughout the paper, we denote scalar values by either lower-case or upper-case characters $(x, y, X, Y \dots)$, vectors by lower-case bold-face characters $(\mathbf{x}, \mathbf{y}, \dots)$, matrices by upper-case or Greek bold-face characters $(\mathbf{A}, \mathbf{B}, \mathbf{\Phi}, \dots)$ and tensor as calligraphic capitals $(\mathcal{X}, \mathcal{Y}, \dots)$. A tensor with $K$ modes and dimension $I_{k}$ in the mode-$k$ is represented as $\mathcal{X} \in \mathbb{R}^{I_1 \times I_2 \times \dots \times I_K}$. The entry in the $i_k$-th index in mode-$k$ for $k=1,\dots, K$ is denoted as $\mathcal{X}_{i_1,i_2,\dots,i_K}$. In addition, $vec(\mathcal{X})$ denotes the vectorization operation that rearranges elements in $\mathcal{X}$ to the vector representation. 

The mode-$k$ product between a tensor $\mathcal{X} \in  \mathbb{R}^{I_1 \times \dots I_K}$ and a matrix $\mathbf{W}\in \mathbb{R}^{J_{k}\times I_k}$ is another tensor of size $I_1\times \dots \times J_{k}\times \dots \times I_K$ and denoted by $\mathcal{X} \times_{k} \mathbf{W}$. The element of $\mathcal{X} \times_{k} \mathbf{W}$ is defined as $[\mathcal{X}\times_{k}\mathbf{W}]_{i_1, \dots , i_{k-1}, j_k, i_{k+1},\dots, i_K}=\sum_{i_k=1}^{I_K}[\mathcal{X}]_{i_1,\dots,i_{k-1},i_k,\dots, i_K}[\mathbf{W}]_{j_k,i_k}$.

In CS, given the original analog signal $\mathbf{y} \in \mathbb{R}^{I}$, signal acquisition model is described via the following equation:

\begin{equation}\label{eq1}
\mathbf{z} = \mathbf{\Phi} \mathbf{y}
\end{equation}
where $\mathbf{z} \in \mathbb{R}^{M}, M \ll I$ denotes the measurements obtained from CS devices and $\mathbf{\Phi}$ denotes the sensing operator or sensing matrix that performs linear interpolation of $\mathbf{y}$. 

For a multidimensional signal $\mathcal{Y} \in \mathbb{R}^{I_1 \times \dots \times I_K}$, we have Multilinear Compressive Sensing (MCS) model as follows:

\begin{equation}\label{eq2}
\mathcal{Z} = \mathcal{Y} \times_1 \mathbf{\Phi}_1 \times \dots \times_K \mathbf{\Phi}_K
\end{equation}
where $\mathcal{Z} \in \mathbb{R}^{M_1 \times \dots \times M_K}, M_k \ll I_k, \forall k=1, \dots, K$ is the compressed measurements having a tensor form, and $\mathbf{\Phi}_k \in \mathbb{R}^{M_k \times I_k}$ denotes separable sensing operators that perform linear transformation along each mode of the original signal $\mathcal{Y}$. 

Since Eq. (\ref{eq2}) can also be written as:

\begin{equation}\label{eq3}
\mathbf{z} = (\mathbf{\Phi}_1 \otimes \dots \otimes \mathbf{\Phi}_K) \mathbf{y}
\end{equation}
where $\mathbf{z} = vec(\mathcal{Z})$, $\mathbf{y} = vec(\mathcal{Y})$, and $\otimes$ denotes the Kronecker product, we can always express MCS in the vector-based fashion, but not vice versa.

\subsection{Compressive Learning}

The objective of Compressive Learning is to create learning models that generate predictions from the compressed measurements $\mathbf{z}$ or $\mathcal{Z}$. This idea was first proposed in the early work of \cite{davenport2007smashed}, where the authors train a classifier directly on the compressed measurements without the signal reconstruction step. It has been shown in this work that when the number of measurements approximates the dimensionality of the data manifold, accurate classifiers can be trained directly in the compressive domain. To achieve good classification performance, an extension of \cite{davenport2007smashed} later showed that the number of measurements only depends on the intrinsic dimensionality of the data manifold \cite{baraniuk2009random}.

A similar theoretical result was derived later in \cite{calderbank2012finding}, which proves that the performance of a linear Support Vector Machine classifier trained on compressed measurements is equivalent to the best linear threshold classifier operating in the original signal domain, given the Distance-Preserving Property holds for the sensing matrices. Asymptotic behavior and sensing operator optimization in CL models of signals described by Gaussian Mixture Model have also been proposed in \cite{reboredo2013compressive, reboredo2013projections}. 

The idea of using past measurements as a prior to jointly learn better sensing matrices and the classifier was proposed in \cite{baheti2008adaptive}. Since the advancement in stochastic optimization, linear classifiers and shallow learning models have been replaced with deep neural networks \cite{lohit2015reconstruction, lohit2016direct}, and separate optimization of the sensing operators and the classifiers have been replaced by the end-to-end training paradigm, which has been shown to significantly outperform predefined sensing operators when the compression rates are high \cite{adler2016compressed, hollis2018compressed, deugerli2018compressively, zisselman2018compressed, tran2019multilinear}. 

\subsection{Multilinear Compressive Learning}

While previous works in learning from compressed measurements adopt the signal acquisition model in Eq. (\ref{eq1}), recently, the authors in \cite{tran2019multilinear} proposed Multilinear Compressive Learning (MCL), a framework that utilizes Eq. (\ref{eq2}) as the signal acquisition model in order to retain the multidimensional structure of the original signal. 

MCL consists of three components: the CS component described by Eq. (\ref{eq2}), the Feature Synthesis (FS) component that transforms the measurements $\mathcal{Z}$ to tensor feature $\mathcal{T}$ via:

\begin{equation}\label{eq4}
\mathcal{T} = \mathcal{Z} \times_1 \mathbf{\Theta}_1 \times \dots \times_K \mathbf{\Theta}_K
\end{equation}
and a task-specific neural network $\mathsf{N}$ that operates on $\mathcal{T}$ to generate prediction, i.e., the prediction is $\mathsf{N}(\mathcal{T})$.

Since the transformation in Eq. (\ref{eq2}) preserves the tensor structure of $\mathcal{Y}$, MCL assumes the existence of a tensor subspace which captures the essential information in $\mathcal{Y}$ through $\mathcal{Z}$, and from which discriminative features can be synthesized. 

Similar to the end-to-end vector-based framework \cite{adler2016compressed, zisselman2018compressed}, MCL incorporates a kind of \textit{weak prior} that initializes $\mathbf{\Phi}_k$ and $\mathbf{\Theta}_k$ with energy-preserving values, i.e., the singular vectors corresponding to the largest singular values in mode-$k$ obtained from HOSVD of the training data. The task-specific network $\mathsf{N}$ is initialized with the weights trained on uncompressed signals. The whole system is then optimized via stochastic gradient descent. As pointed out in the empirical analysis in \cite{tran2019multilinear}, although simple, this initialization scheme contributes significantly to the performances of the system compared to the de facto random initialization scheme often utilized in deep neural networks \cite{he2016identity}. 

\subsection{Knowledge Distillation}

The term Knowledge Distillation (KD) was coined in \cite{hinton2015distilling}, in which the authors proposed a neural network training technique that utilizes the prediction from a pre-trained high capacity network (the teacher network) to provide supervisory signals to a smaller network (the student network) along with labeled data. The intuition behind this training technique is that with higher capacity, it is easier for the teacher network to discover better data representation, and the form of knowledge provided via the teacher network's prediction helps guide the student network to better solutions.  

There have been several works investigating variants of this technique. For example, KD that involves multiple student and teacher networks \cite{zhang2018deep} has been shown to be more robust than a single teacher-student pair. In other works \cite{romero2014fitnets,yim2017gift,yu2017visual}, in addition to the predicted probabilities, knowledge coming from intermediate layers of the teacher network has also been proven to be useful for the student. While KD has often been considered in the context of model compression, i.e., to train low capacity models with better performances, this paradigm has also been successfully applied to distributed training with student and teacher networks having the same architecture \cite{anil2018large}. 

In our work, we propose to use KD as a method to progressively incorporate prior information into CL models. While in a general setting it is unclear how to select the immediate source and target layers to transfer knowledge from the teacher to the student, we will show next that in the context of CL, and especially in the MCL framework, pairing the intermediate teachers and students is easy since the learning system is modularized into different components with different functionalities.

\section{Proposed Methods}\label{S3}

\subsection{Finding Prior Knowledge}\label{S3.1}

\begin{figure*}[]
	\centering
	\includegraphics[width=0.95\linewidth]{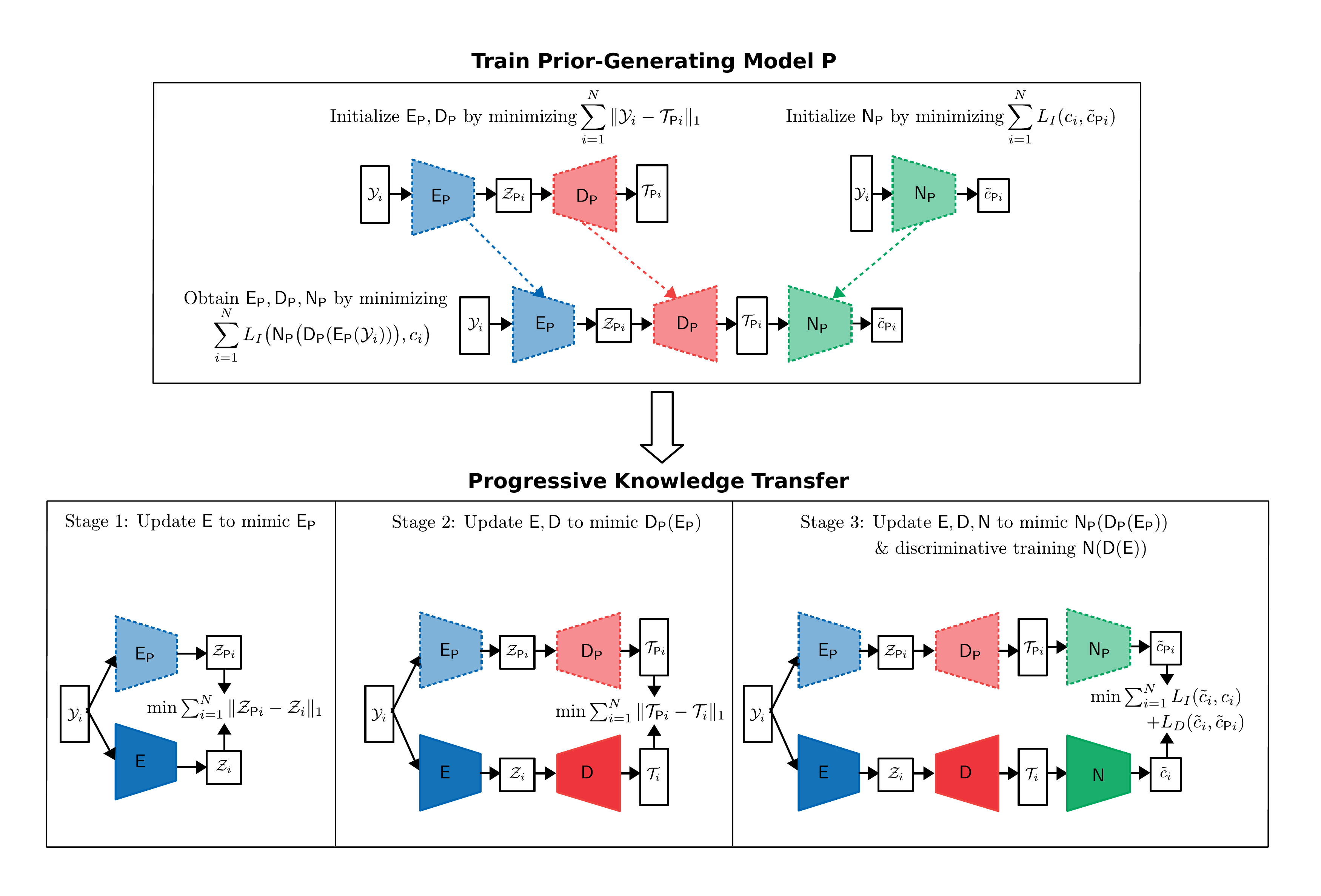}
	\caption{Illustration of the proposed algorithm: we first train prior-generating model $\mathsf{P}$. The knowledge in each component of $\mathsf{P}$ is then progressively transferred to the MCL model. Lighter color blocks indicate components of $\mathsf{P}$ while darker color ones indicate the corresponding components in MCL model.}
	\label{f1}
\end{figure*} 

A common constraint which exists in all CL models is that the sensing operator is linear or multilinear. This limits the capacity of CL models and the amount of information that can be captured from the original signal. Thus, in CL in general and MCL in particular, given a fixed structure of the Feature Synthesis (FS) component and the task-specific neural network $\mathsf{N}$, there is always an upper-bound on the capacity of the whole learning system, which makes the problem of learning from compressed measurements $\mathcal{Z}$ generally harder than other unconstrained learning tasks.   

As mentioned in the Introduction Section, finding and incorporating prior knowledge is a recurring theme in CS community since good priors have been shown both theoretically and empirically to improve signal recovery and compression efficiency \cite{mota2017compressed}. In case of CL, prior knowledge is less obvious. However, we may still formulate the following question: ´´\textit{What kind of prior knowledge do we know about the measurements $\mathcal{Z}$ and the types of features $\mathcal{T}$ that are representative for the learning task?}"

Although we might not have direct knowledge of optimal $\mathcal{Z}$ or $\mathcal{T}$, we know as a prior knowledge that, given sufficient training data, a higher capacity neural network is expected to perform better than a less complex one when both networks are structurally similar. By structure here we mean that both networks have similar connectivity patterns and by higher capacity, we mean higher number of parameters and/or higher number of layers. The structural similarity ensures a feasible capacity comparison between networks since one cannot make such a guess about the learning ability, e.g., between a feed-forward architecture having a higher parameter count and a residual architecture having a lower parameter count. Therefore, we almost know for certain that given an MCL model with upper-bounded learning capacity, one can construct and train other learning systems with similar FS component and task-specific network $\mathsf{N}$, but having nonlinear sensing operators to achieve better learning performance, thus better $\mathcal{Z}$ and $\mathcal{T}$. The representations produced by higher capacity, nonlinear compressive learning models become our direct prior knowledge of the compressive domain and the feature space. 

The advantage of the above presented approach to define prior knowledge in MCL is two-fold:

\begin{itemize}
\item Since the priors are generated from another learning system trained on the same learning task, we incorporate completely data-dependent, task-specific priors into our model instead of hand-crafted priors. 

\item By sharing the same structure of the FS component and task-specific neural network $\mathsf{N}$ between the MCL model and the prior-generating model, the weights obtained from the latter provide a good initialization when optimizing the former. 
\end{itemize}

The proposed approach reduces the problem of finding priors to the task of designing and training a nonlinear compressive learning model; hereupon we refer to it as the prior-generating model $\mathsf{P}$, which has no structural constraint. To construct and optimize $\mathsf{P}$, we propose to mirror the strategy in MCL \cite{tran2019multilinear}: 

\begin{itemize}
\item The nonlinear sensing component of $\mathsf{P}$, denoted as $\mathsf{E}_{\mathsf{P}}$ with parameters $\theta_{\mathsf{E_P}}$, has several layers to reduce the dimensionality of the original signal gradually. $\mathsf{E}_{\mathsf{P}}$ transforms the input signal $\mathcal{Y}$ to a compressed representation $\mathcal{Z}_{\mathsf{P}}$, i.e., $\mathcal{Z}_{\mathsf{P}} = \mathsf{E}_{\mathsf{P}}(\mathcal{Y}; \theta_{\mathsf{E_P}})$, which has the same size as $\mathcal{Z}$. 

\item Similarly, the nonlinear FS component of $\mathsf{P}$, denoted as $\mathsf{D}_{\mathsf{P}}$ with parameters $\theta_{\mathsf{D_P}}$, has several layers to increase the dimensionality of $\mathcal{Z}_{\mathsf{P}}$ gradually, producing features $\mathcal{T}_{\mathsf{P}} = \mathsf{D_P}(\mathcal{Z}_{\mathsf{P}}; \theta_{\mathsf{D_P}})$, which has the same dimensionality as the original signal. $\mathsf{E}_{\mathsf{P}}$ and $\mathsf{D}_{\mathsf{P}}$ together resemble the encoder and decoder of a nonlinear autoencoder. 

\item Denote $\mathcal{Y}_i$ and $c_i$ ($i=1, \dots, N$) the $i$-th training sample and its label. In order to initialize the weights $\theta_{\mathsf{E}_P}$ and $\theta_{\mathsf{D_P}}$, we update $\mathsf{E_P}$ and $\mathsf{D_P}$ to minimize $\sum_{i=1}^{N} \| \mathcal{Y}_i - \mathsf{D}_{\mathsf{P}}\big(\mathsf{E}_{\mathsf{P}}(\mathcal{Y}_i)\big) \|_1$ via stochastic gradient descent. The weights $\theta_{\mathsf{N_P}}$ of task-specific network $\mathsf{N}_{\mathsf{P}}$ are initialized with weights minimizing $\sum_{i=1}^{N} L_I(\mathsf{N}_{\mathsf{P}}(\mathcal{Y}_i), c_i)$ with $L_I(\cdot)$ is the inference loss. 

\item After the initialization step, $\theta_{\mathsf{E}_{\mathsf{P}}}$, $\theta_{\mathsf{D}_{\mathsf{P}}}$, and $\theta_{\mathsf{N}_{\mathsf{P}}}$ are updated altogether via stochastic gradient descent to minimize $\sum_{i=1}^{N} L_I \big(\mathsf{N}_{\mathsf{P}}\big(\mathsf{D}_{\mathsf{P}}(\mathsf{E}_{\mathsf{P}}(\mathcal{Y}_i))\big), c_i\big)$
\end{itemize}

\begin{algorithm}
	\caption{Training MCL Model via Knowledge Transfer Algorithm in Supervised Learning Setting}
	\label{algo1}
	\begin{algorithmic}[1]
		\State \textbf{Inputs}: 
		\State \hskip1.5em Labeled data set $\mathfrak{L}=\{(\mathcal{Y}_i, c_i) | i=1,\dots, N\}$.
		\State \hskip1.5em Structure of prior-generating model $\mathsf{E_P}$, $\mathsf{D_P}$, $\mathsf{N_P}$
		\State \hskip1.5em Structure of MCL model $\mathsf{E}$, $\mathsf{D}$, $\mathsf{N}$
		\State \hskip1.5em Inference loss $L_I$, Distillation loss $L_D$  
		
		\State \textbf{Training prior-generating model $\mathsf{P}$}:
		\State \hskip1.5em Initialize $\theta_{\mathsf{N_P}}$ with values from 
		\State \hskip3.0em $\displaystyle \argmin_{\theta_{\mathsf{N_P}}}\sum_{i=1}^{N} L_I(\mathsf{N}_{\mathsf{P}}(\mathcal{Y}_i), c_i)$
		\State \hskip1.5em Initialize $\theta_{\mathsf{E_P}}$ and $\theta_{\mathsf{D_P}}$ with values from 
		\State \hskip3.0em $\displaystyle \argmin_{\theta_{\mathsf{E_P}}, \theta_{\mathsf{D_P}}} \sum_{i=1}^{N} \| \mathcal{Y}_i - \mathsf{D}_{\mathsf{P}}\big(\mathsf{E}_{\mathsf{P}}(\mathcal{Y}_i)\big) \|_1$
		\State \hskip1.5em Obtain $\theta_{\mathsf{E_P}}$, $\theta_{\mathsf{D_P}}$, $\theta_{\mathsf{N_P}}$ by solving
		\State \hskip3.0em $\displaystyle \argmin_{\theta_{\mathsf{E_P}}, \theta_{\mathsf{D_P}}, \theta_{\mathsf{N_P}}} \sum_{i=1}^{N} L_I \big(\mathsf{N}_{\mathsf{P}}\big(\mathsf{D}_{\mathsf{P}}(\mathsf{E}_{\mathsf{P}}(\mathcal{Y}_i))\big), c_i\big)$
		\State \textbf{Training MCL model}:
		\State \hskip1.5em After obtaining $\theta_{\mathsf{E_P}}$, $\theta_{\mathsf{D_P}}$, $\theta_{\mathsf{N_P}}$, they are fixed.
		\State \hskip1.5em Transfer sensing knowledge:
		\State \hskip3.0em Update $\theta_{\mathsf{E}}$ by solving Eq. (\ref{eq5})
		\State \hskip1.5em Transfer feature synthesis knowledge: 
		\State \hskip3.0em Update $\theta_{\mathsf{E}}$, $\theta_{\mathsf{D}}$  by solving Eq. (\ref{eq6})
		\State \hskip1.5em Transfer inference knowledge \& discriminative training:
		\State \hskip3.0em Update $\theta_{\mathsf{E}}$, $\theta_{\mathsf{D}}$, $\theta_{\mathsf{N}}$ by solving Eq. (\ref{eq7})
		\State \textbf{Outputs}: Parameters of MCL model: $\theta_{\mathsf{E}}$, $\theta_{\mathsf{D}}$, $\theta_{\mathsf{N}}$
	\end{algorithmic}
\end{algorithm}

\subsection{Incorporating Prior Knowledge}\label{S3.2}

Let us denote the CS, FS and task-specific neural network component of the MCL model as $\mathsf{E}$, $\mathsf{D}$ and $\mathsf{N}$, having learnable parameters $\theta_{\mathsf{E}}$, $\theta_{\mathsf{D}}$ and $\theta_{\mathsf{N}}$, respectively. It should be noted that $\mathsf{D}$ and $\mathsf{N}$ have the same functional form as $\mathsf{D_P}$ and $\mathsf{N_P}$ while $\mathsf{E}$ possesses a multilinear form as in Eq. (\ref{eq2}).

After training the prior-generating model $\mathsf{P}$, we perform progressive KD to transfer the knowledge of $\mathsf{P}$ to the MCL model. During this phase, the parameters of $\mathsf{P}$, i.e., $\theta_{\mathsf{E_P}}$, $\theta_{\mathsf{D_P}}$, and $\theta_{\mathsf{N_P}}$ are fixed. Since the teacher ($\mathsf{P}$) and the student (MCL) model share the same modular structure which has three distinct components, training MCL with prior knowledge given by $\mathsf{P}$ consists of three stages:

\begin{itemize}
\item \textbf{Stage 1} (Transferring the sensing knowledge from $\mathsf{E_P}$ to $\mathsf{E}$): during this stage, the weights $\theta_{\mathsf{E}}$ of the sensing component $\mathsf{E}$ in MCL are obtained by optimizing the following criterion:

\begin{equation}\label{eq5}
\argmin_{\theta_{\mathsf{E}}} \sum_{i=1}^{N} \| \mathsf{E_P}(\mathcal{Y}_i; \theta_{\mathsf{E_P}}) - \mathsf{E}(\mathcal{Y}_i ; \theta_{\mathsf{E}}) \|_1
\end{equation}
The purpose of this stage is to enforce the sensing component in MCL to mimic that of the prior-generating model $\mathsf{P}$.

\item \textbf{Stage 2} (Transferring the feature synthesis knowledge from $\mathsf{D_P}(\mathsf{E_P})$ to $\mathsf{D}(\mathsf{E})$): during this stage, we aim to optimize $\mathsf{E}$ and $\mathsf{D}$ in MCL to synthesize similar features produced by the prior-generating model. Before the optimization process, the weights of the sensing component ($\theta_{\mathsf{E}}$) are initialized with values obtained from the 1\textsuperscript{st} stage, and the weights of the feature synthesis component $\theta_{\mathsf{D}}$ are initialized with values from $\theta_{\mathsf{D_P}}$ since $\mathsf{D}$ in MCL and $\mathsf{D_P}$ in $\mathsf{P}$ have the same functional form. After the initialization step, both sensing ($\theta_{\mathsf{E}}$) and feature synthesis ($\theta_{\mathsf{D}}$) components in MCL are updated together to minimize the following criterion:

\begin{equation}\label{eq6}
\argmin_{\theta_{\mathsf{E}}, \theta_{\mathsf{D}}} \sum_{i=1}^{N} \| \mathsf{D_P}\big(\mathsf{E_P}(\mathcal{Y}_i; \theta_{\mathsf{E_P}}); \theta_{\mathsf{D_P}} \big) - \mathsf{D}\big(\mathsf{E}(\mathcal{Y}_i; \theta_{\mathsf{E}}); \theta_{\mathsf{D}} \big) \|_1
\end{equation}

\item \textbf{Stage 3} (Transferring the inference knowledge from $\mathsf{N_P}(\mathsf{D_P}(\mathsf{E_P}))$ to $\mathsf{N}(\mathsf{D}(\mathsf{E}))$ and discriminative training $\mathsf{N}(\mathsf{D}(\mathsf{E}))$): in this final stage, the MCL model is trained to minimize the inference loss as well as the difference between its prediction and the prediction produced by the prior-generating model $\mathsf{P}$. That is, the minimization objective in this stage is:

\begin{equation}\label{eq7}
\begin{split}
&\argmin_{\theta_{\mathsf{E}}, \theta_{\mathsf{D}}, \theta_{\mathsf{N}}}  
\sum_{i=1}^{N} L_I \big(\mathsf{N}\big(\mathsf{D}(\mathsf{E}(\mathcal{Y}_i))\big), c_i\big) + \\
&\lambda L_D\big( \mathsf{N_P}\big(\mathsf{D_P}\big(\mathsf{E_P}(\mathcal{Y}_i)\big)\big),  \mathsf{N}\big(\mathsf{D}\big(\mathsf{E}(\mathcal{Y}_i)\big)\big)\big)
\end{split}
\end{equation}
where $L_I(\cdot)$ and $L_D(\cdot)$ denote the inference loss and distillation loss, respectively. $\lambda$ is a hyper-parameter that allows the adjustment of distillation loss. The specific form of $L_I(\cdot)$ and $L_D(\cdot)$ are chosen depending on the inference problem. For example, in classification problems, we can select cross-entropy function for $L_I(\cdot)$ and Kullback-Leibler divergence for $L_D(\cdot)$ while in regression problems, $L_I(\cdot)$ and $L_D(\cdot)$ can be mean-squared-error or mean-absolute-error function. To avoid lengthy representation, Eq. (\ref{eq7}) omits the parameters of each component. Here we should note that before optimizing Eq. (\ref{eq7}), parameters $\theta_{\mathsf{E}}$ and  $\theta_{\mathsf{D}}$ of the sensing and synthesis component in MCL are initialized with values obtained from the previous stage, while parameters of the task-dependent neural network $\theta_{\mathsf{N}}$ in MCL is initialized with values from $\theta_{\mathsf{N_P}}$ in the prior-generating model $\mathsf{P}$. 
\end{itemize}

The pseudo-code and the illustration of our proposed algorithm to train MCL model via progressive knowledge transfer is presented in Algorithm \ref{algo1} and Figure \ref{f1}, respectively. 

To use KD and the prior-generating model $\mathsf{P}$, a simpler and more straightforward approach is to directly distill the predictions of $\mathsf{P}$ to the MCL model as in the 3\textsuperscript{rd} stage described above, skipping the 1\textsuperscript{st} and 2\textsuperscript{nd} stages. However, by gradually transferring knowledge from each component of the prior-generating model $\mathsf{P}$ to the MCL model, we argue that the proposed training scheme directly incorporates the knowledge of a good compressive domain and feature space, facilitating the MCL model to mimic the internal representations of its teacher, which is better than simply imitating its teacher's predictions. To validate our argument, we provide empirical analysis of the effects of each training stage in Section \ref{S4.6}. 

The aforementioned progressive knowledge transfer scheme also allows us to directly use the predictions produced by the teacher model in the 3\textsuperscript{rd} stage, instead of the ``\textit{softened}" predictions as in the original work \cite{hinton2015distilling}, eliminating the need to select the associated temperature hyper-parameter to soften the teacher's predictions.

\subsection{Semi-supervised Learning Extension}\label{S3.3}

It should be noted that without sufficient labeled data, training a high-capacity prior-generating model as proposed in Section \ref{S3.1} can lead to an over-fitted teacher, which might provide misleading supervisory information to the MCL model. In addition, a limited amount of data might also prevent effective knowledge transfer from the teacher to the student model. In certain scenarios, labeled data is scarce, however, we can easily obtain a large amount of data coming from the same distribution without labels. Semi-supervised learning refers to the learning paradigm that takes advantages of unlabeled data, usually available in abundance, along with a limited amount of labeled data. Here we also describe a semi-supervised adaptation for the above MCL model training technique to remedy possible over-fitting cases and improve generalization of both prior-generating and MCL models in classification tasks. To this end, unlabeled data is utilized to both initialize, and then optimize the weights of the prior-generating model $\mathsf{P}$ via an incremental self-labeling procedure. Subsequently, when transferring knowledge to the MCL model, the class predictions of $\mathsf{P}$ on unlabeled data are used as \textit{hard} labels in the inference loss term, while its probability predictions on unlabeled data are used as \textit{soft} labels in the distillation loss term.  

Let us denote $\mathfrak{L} = \{(\mathcal{Y}_i, c_i) | i=1, \dots, N\}$ the labeled training set and $\mathfrak{U} = \{ \mathcal{Y}_i | i=N+1, \dots, M\}$ the unlabeled training set. 
To take advatange of $\mathfrak{L} \cup \mathfrak{U}$, we propose the following modifications to the training procedure of prior-generating model $\mathsf{P}$:

\begin{itemize}
\item Initialization of $\mathsf{E_P}$ and $\mathsf{D_P}$: the weights of the sensing ($\theta_{\mathsf{E_P}}$) and feature synthesis ($\theta_{\mathsf{D_P}}$) component are initialized with values obtained from minimizing the reconstruction error on $\mathfrak{L} \cup \mathfrak{U}$, i.e., $\sum_{i=1}^{M} \| \mathcal{Y}_i - \mathsf{D}_{\mathsf{P}}\big(\mathsf{E}_{\mathsf{P}}(\mathcal{Y}_i)\big) \|_1$. 

\item Incremental optimization of $\mathsf{E_P}, \mathsf{D_P}, \mathsf{N_P}$ via self-labeling: after the initialization step, all parameters of the prior-generating model $\mathsf{P}$ are optimized with respect to the inference loss, which is calculated on the enlarged labeled set $\tilde{\mathfrak{L}}$:

\begin{equation}\label{eq7.1}
\argmin_{\theta_{\mathsf{E_P}}, \theta_{\mathsf{D_P}}, \theta_{\mathsf{N_P}}} \sum_{(\mathcal{Y}_i, c_i) \in \tilde{\mathfrak{L}}} L_I \big(\mathsf{N}_{\mathsf{P}}\big(\mathsf{D}_{\mathsf{P}}(\mathsf{E}_{\mathsf{P}}(\mathcal{Y}_i))\big), c_i\big)
\end{equation}

Initially, the enlarged labeled set is formed from the labeled data, i.e., $\tilde{\mathfrak{L}} = \mathfrak{L}$. After every $T$ backpropagation epochs, $\tilde{\mathfrak{L}}$ is augmented with those data instances (with their predicted labels) in $\mathfrak{U}$ that have the most confident predictions from the current $\mathsf{P}$, given a probability threshold $\rho$, i.e., $\tilde{\mathfrak{L}} = \tilde{\mathfrak{L}} \cup \mathfrak{C}$:

\begin{equation}\label{eq7.2}
\begin{split}
\mathfrak{C} =  \{ &(\mathcal{Y},c) \, | \,\mathcal{Y} \in \mathfrak{U} \:\wedge \: \mathsf{N_P}(\mathsf{D_P(\mathsf{E_P}(\mathcal{Y}))})_{\max} \geq \rho, \\
& c=\textrm{argmax}(\mathsf{N_P}(\mathsf{D_P(\mathsf{E_P}(\mathcal{Y}))}))\}
\end{split}
\end{equation}

After the enlargement of $\tilde{\mathfrak{L}}$ with the most confident instances, they are removed from the unlabeled set $\mathfrak{U}$, i.e., $\mathfrak{U} = \mathfrak{U} \setminus \mathfrak{C}$. The training terminates when the enlargement of $\tilde{\mathfrak{L}}$ stops, i.e., $\mathfrak{C} = \varnothing$.

\end{itemize}

Self-labeling is a popular technique in semi-supervised algorithms. While there are many sophisticated variants of this technique \cite{triguero2015self}, the simple modifications proposed above work well as illustrated in our empirical study in Section \ref{S4.7}. 

Given a prior-generating model $\mathsf{P}$ trained on $\mathfrak{L} \cup \mathfrak{U}$, in order to adapt the progressive knowledge transfer algorithm proposed in Section \ref{S3.2} to the semi-supervised setting, we propose to replace the objectives in Eq. (\ref{eq5}), (\ref{eq6}) and (\ref{eq7}) with the following objectives respectively:

\begin{equation}\label{eq8}
\argmin_{\theta_{\mathsf{E}}} \sum_{i=1}^{M} \| \mathsf{E_P}(\mathcal{Y}_i; \theta_{\mathsf{E_P}}) - \mathsf{E}(\mathcal{Y}_i ; \theta_{\mathsf{E}}) \|_1
\end{equation}
and
\begin{equation}\label{eq9}
\argmin_{\theta_{\mathsf{E}}, \theta_{\mathsf{D}}} \sum_{i=1}^{M} \| \mathsf{D_P}\big(\mathsf{E_P}(\mathcal{Y}_i; \theta_{\mathsf{E_P}}); \theta_{\mathsf{D_P}} \big) - \mathsf{D}\big(\mathsf{E}(\mathcal{Y}_i; \theta_{\mathsf{E}}); \theta_{\mathsf{D}} \big) \|_1
\end{equation}
and
\begin{equation}\label{eq10}
\begin{split}
&\argmin_{\theta_{\mathsf{E}}, \theta_{\mathsf{D}}, \theta_{\mathsf{N}}}  
\sum_{i=1}^{M} L_I \big(\mathsf{N}\big(\mathsf{D}(\mathsf{E}(\mathcal{Y}_i))\big), c_i\big) + \\
&\lambda L_D\big( \mathsf{N_P}\big(\mathsf{D_P}\big(\mathsf{E_P}(\mathcal{Y}_i)\big)\big),  \mathsf{N}\big(\mathsf{D}\big(\mathsf{E}(\mathcal{Y}_i)\big)\big)\big)
\end{split}
\end{equation}
where in (\ref{eq10}), $c_i$ denotes the class label predicted by the prior-generating model $\mathsf{P}$ for $i=N+1, \dots, M$. 
 
A summary of our proposed algorithm in the semi-supervised setting is presented in Algorithm \ref{algo2}.

\begin{algorithm}
	\caption{Training MCL Model via Knowledge Transfer Algorithm in Semi-supervised Learning Setting}
	\label{algo2}
	\begin{algorithmic}[1]
		\State \textbf{Inputs}: 
		\State \hskip1.5em Labeled data set $\mathfrak{L}=\{(\mathcal{Y}_i, c_i) | i=1,\dots, N\}$.
		\State \hskip1.5em Unlabeled data set $\mathfrak{U}=\{\mathcal{Y}_i | i=N+1,\dots, M\}$.
		\State \hskip1.5em Structure of prior-generating model $\mathsf{E_P}$, $\mathsf{D_P}$, $\mathsf{N_P}$
		\State \hskip1.5em Structure of MCL model $\mathsf{E}$, $\mathsf{D}$, $\mathsf{N}$
		\State \hskip1.5em Inference loss $L_I$, Distillation loss $L_D$  
		\State \hskip1.5em Number of backpropagation iteration $T$
		\State \hskip1.5em Confidence probability threshold $\rho$
		
		\State \textbf{Training prior-generating model $\mathsf{P}$}:
		\State \hskip1.5em Initialize $\theta_{\mathsf{N_P}}$ with values from 
		\State \hskip3.0em $\displaystyle \argmin_{\theta_{\mathsf{N_P}}}\sum_{i=1}^{N} L_I(\mathsf{N}_{\mathsf{P}}(\mathcal{Y}_i), c_i)$
		\State \hskip1.5em Initialize $\theta_{\mathsf{E_P}}$ and $\theta_{\mathsf{D_P}}$ with values from 
		\State \hskip3.0em $\displaystyle \argmin_{\theta_{\mathsf{E_P}}, \theta_{\mathsf{D_P}}} \sum_{i=1}^{M} \| \mathcal{Y}_i - \mathsf{D}_{\mathsf{P}}\big(\mathsf{E}_{\mathsf{P}}(\mathcal{Y}_i)\big) \|_1$
		\State \hskip1.5em Let $\tilde{\mathfrak{L}} = \mathfrak{L}$
		\State \hskip1.5em \textbf{while} True \textbf{do}
		\State \hskip3.0em Update $\theta_{\mathsf{E_P}}$, $\theta_{\mathsf{D_P}}$, $\theta_{\mathsf{N_P}}$ by solving Eq. (\ref{eq7.1}) for $T$ iterations
		\State \hskip3.0em Find set of confident instances according to Eq. (\ref{eq7.2})
		\State \hskip3.0em \textbf{if} $\mathfrak{C} = \varnothing$ \textbf{then}
		\State \hskip4.0em \textbf{break}
		\State \hskip3.0em Augment enlarged labeled set $\tilde{\mathfrak{L}} = \tilde{\mathfrak{L}} \cup \mathfrak{C}$
		\State \hskip3.0em Reduce unlabeled set $\mathfrak{U} = \mathfrak{U} \setminus \mathfrak{C}$
		\State \textbf{Training MCL model}:
		\State \hskip1.5em After obtaining $\theta_{\mathsf{E_P}}$, $\theta_{\mathsf{D_P}}$, $\theta_{\mathsf{N_P}}$, they are fixed.
		\State \hskip1.5em Transfer sensing knowledge:
		\State \hskip3.0em Update $\theta_{\mathsf{E}}$ by solving Eq. (\ref{eq8})
		\State \hskip1.5em Transfer feature synthesis knowledge: 
		\State \hskip3.0em Update $\theta_{\mathsf{E}}$, $\theta_{\mathsf{D}}$  by solving Eq. (\ref{eq9})
		\State \hskip1.5em Transfer inference knowledge \& discriminative training:
		\State \hskip3.0em Update $\theta_{\mathsf{E}}$, $\theta_{\mathsf{D}}$, $\theta_{\mathsf{N}}$ by solving Eq. (\ref{eq10})
		\State \textbf{Outputs}: Parameters of MCL model: $\theta_{\mathsf{E}}$, $\theta_{\mathsf{D}}$, $\theta_{\mathsf{N}}$
	\end{algorithmic}
\end{algorithm}

\section{Experiments}\label{S4}
In this Section, we provide a detailed description and results of our empirical analysis which demonstrates the advantages of the proposed algorithms that incorporate data-dependent prior knowledge into the training procedure of MCL models, compared to \cite{tran2019multilinear}. For this purpose, we provide various comparisons between the proposed algorithm and the original algorithm proposed in \cite{tran2019multilinear} in the supervised learning setting. The experiments are designed to benchmark the learning performances and the quality of signal representation in compressive domain produced by the two competing algorithms, with increasing difficulty in the learning tasks. In addition, we also study the significance of prior-generating model and different knowledge transfer stages via different ablation experiments. Lastly, we demonstrate the necessity of the proposed modifications presented in Section \ref{S3.3} to our algorithm in the semi-supervised learning setting, i.e., when the databases are large, but only limited amounts of labeled data exist. 

\subsection{Datasets}

We conducted experiments on the object classification and face recognition tasks of the following datasets:

\begin{itemize}
	\item CIFAR dataset \cite{krizhevsky2009learning} is a color (RGB) image dataset used for evaluating object recognition methods. The dataset consists of $50\mathrm{K}$ images for training and $10\mathrm{K}$ images for testing with resolution $32\times 32$ pixels. In our work, CIFAR-10 refers to the $10$-class objection recognition task in which each individual image has a single class label coming from $10$ different categories. A more fine-grained and difficult classification task also exists in CIFAR dataset with each image having a label coming from $100$ different categories, which we denote as CIFAR-100. Since there is no validation set in the original database, in our experiments, we randomly selected $5\mathrm{K}$ images from the training set of CIFAR-10 and CIFAR-100 for validation and only trained the algorithms on $45\mathrm{K}$ images. 
	
	\item CelebA: CelebA \cite{liu2015faceattributes} is a large-scale human face image dataset with more than $200\mathrm{K}$ images at different resolutions from more than $10\mathrm{K}$ identities. In our experiment, we created three versions of CelebA with increasing difficulties by increasing the set of identities to be recognized: CelebA-100, CelebA-200, and CelebA-500 having $100$, $200$, and $500$ identities respectively. Here we note that CelebA-100 is a subset of CelebA-200, and CelebA-200 is a subset of CelebA-500. 
	
	\item To study the performances of the proposed algorithm in semi-supervised settings, we also created CIFAR-10S, CIFAR-100S, CelebA-500S, which have the same number of training instances as CIFAR-10, CIFAR-100, and CelebA-500, respectively, but only 20\% of them are labeled. The test sets in CIFAR-10S, CIFAR-100S and CelebA-500S are the same as those in CIFAR-10, CIFAR-100, and CelebA-500. 
	
\end{itemize}

The information of all datasets used in our experiments is summarized in Table \ref{t1}.

\begin{table*}[t!]
	\begin{center}
		\caption{Datasets Information}\label{t1}
		\resizebox{0.98\textwidth}{!}{
			\begin{tabular}{|c|c|c|c|c|c|c|c|}\hline
Dataset 	& Input Dimension & Output Dimension	& \#Labeled Train	& \#Unlabeled Train & \#Validation	& \#Test 
\\ \hline
CIFAR-10	& $32\times 32\times 3$ & $10$  & $45000$ & $0$	& $5000$ & $10000$ \\ \hline
CIFAR-100	& $32\times 32\times 3$ & $100$  & $45000$ & $0$	& $5000$ & $10000$ \\ \hline
CelebA-100	& $32\times 32\times 3$ & $100$  & $7063$ & $0$	& $2373$ & $2400$ \\ \hline
CelebA-200	& $32\times 32\times 3$ & $200$  & $14305$ & $0$	& $4807$ & $4860$ \\ \hline
CelebA-500	& $32\times 32\times 3$ & $500$  & $35805$ & $0$	& $12026$ & $12169$ \\ \hline \hline
CIFAR-10S	& $32\times 32\times 3$ & $10$  & $9000$ & $36000$	& $1500$ & $10000$ \\ \hline
CIFAR-100S	& $32\times 32\times 3$ & $100$  & $9000$ & $36000$	& $1500$ & $10000$ \\ \hline
CelebA-500S	& $32\times 32\times 3$ & $500$  & $7009$ & $28796$	& $3431$ & $12169$ \\ \hline

			\end{tabular}
		}
	\end{center}
\end{table*}

\subsection{Experiment Protocols}

In this Section, to differentiate between MCL models trained by different algorithms, we use the abbreviation MCL to refer to the original algorithm proposed in \cite{tran2019multilinear}, MCLwP to refer to our proposed algorithm that trains MCL models with a prior-generating model with labeled data only, and MCLwP-S to refer to our proposed algorithm that can take advantages of additional unlabeled data. 

Regarding the architectural choices in MCL, MCLwP, MCLwP-S, we adopted the AllCNN architecture \cite{tran2019multilinear} for the task-specific neural network component in all algorithms. In MCL, the CS and FS components both perform multilinear transformation while in MCLwP and MCLwP-S, the CS component performs multilinear transformation and the FS component performs nonlinear transformation which consists of both convolution and multilinear projection. The exact network architecture used for each algorithm can be found in our publicly available implementation\footnote{https://github.com/viebboy/MultilinearCompressiveLearningWithPrior}. 

To compare the learning performances at different measurement rates, we have conducted experiments at four measurement rates $0.01$, $0.02$, $0.05$ and $0.10$, corresponding to the following configurations of $\mathcal{Z}$: $6\times 6\times 1$, $9\times 7\times 1$, $13\times 12\times 1$, and $14\times 11\times 2$, respectively. 

Since all learning tasks are classification tasks, we represented the labels with one-hot encoding and used cross-entropy function and symmetric Kullback-Leibler divergence for the inference loss $L_I$ and distillation loss $L_D$, respectively. For each experiment configuration, we performed three runs, and the median values of accuracy on the test set are reported. For $\lambda$, the hyper-parameter that controls the amount of distillation loss in Eq. (\ref{eq7}), we set it equal to $1.0$. Confidence probability threshold $\rho$ in semi-supervised learning experiments was selected from the set $\{0.7, 0.8, 0.9\}$. 

Regarding stochastic optimization protocol, we used ADAM optimizer \cite{kingma2014adam} with the following learning rate schedule $\{10^{-3}, 10^{-4}, 10^{-5}\}$, changing at epochs $80$ and $120$. Each stochastic optimization procedure was conducted for $160$ epochs in total. Max-norm constraint with a value of $6.0$ was used to regularize the parameters in all networks. No data preprocessing was conducted, except the scaling of all pixel values to $[0, 1]$. During stochastic optimization, we performed random flipping on the horizontal axis and image shifting within $10\%$ of the spatial dimensions to augment the training set. In all experiments, the final model weights, which are used to measure the performance on the test sets, are obtained from the epoch, which has the highest validation accuracy.

\subsection{Comparisons with MCL \cite{tran2019multilinear}}\label{S4.3}
%
%
%
\begin{table*}[t!]
	\begin{center}
		\caption{Learning performances of MCL \cite{tran2019multilinear} and MCLwP (our proposed algorithm) measured on test set (accuracy in \%). The last four rows show the performances of prior-generating model $\mathsf{P}$ for reference}\label{t2}
		\resizebox{0.85\textwidth}{!}{
\begin{tabular}{|c|c|c|c|c|c|c|} \hline
	Measurements           & Models & CIFAR-10 & CIFAR-100 & CelebA-100 & CelebA-200 & CelebA-500 \\ \hline \hline
	\multirow{3}{*}{$6\times 6\times 1$} & MCL \cite{tran2019multilinear}  %
	& $\mathbf{58.43}$    & $27.55$     & $53.96$      & $44.52$      & $38.07$      \\ \cline{2-7}
	& MCLwP (our)  %
	& $57.80$    & $\mathbf{31.17}$     & $\mathbf{55.31}$      & $\mathbf{47.61}$      & $\mathbf{42.51}$      \\ \cline{2-7}
	& $\Delta$ (MCLwP$-$MCL) & $\texttt{-}0.63$ & $\texttt{+}3.62$ & $\texttt{+}1.35$ & $\texttt{+}3.09$ & $\texttt{+}4.44$ \\ \hline \hline
\multirow{3}{*}{$9\times 7\times 1$} & MCL \cite{tran2019multilinear}    %
& $66.14$    & $33.70$     & $72.08$      & $67.40$      & $61.53$      \\ \cline{2-7}
& MCLwP (our)  %
& $\mathbf{66.81}$    & $\mathbf{36.60}$     & $\mathbf{75.50}$      & $\mathbf{72.02}$      & $\mathbf{68.73}$      \\ \cline{2-7}
	& $\Delta$ (MCLwP$-$MCL) & $\texttt{+}0.67$ & $\texttt{+}2.90$ & $\texttt{+}3.42$ & $\texttt{+}4.62$ & $\texttt{+}7.20$ \\ \hline \hline
\multirow{3}{*}{$13\times 12\times 1$} & MCL \cite{tran2019multilinear}    %
& $77.17$  & $43.49$   & $86.27$    & $83.54$    & $83.63$    \\   \cline{2-7}
& MCLwP (our) %
& $\mathbf{77.58}$  & $\mathbf{47.92}$   & $\mathbf{87.35}$    & $\mathbf{86.39}$    & $\mathbf{85.97}$    \\ \cline{2-7}
	& $\Delta$ (MCLwP$-$MCL) & $\texttt{+}0.41$ & $\texttt{+}4.43$ & $\texttt{+}1.08$ & $\texttt{+}2.85$ & $\texttt{+}2.34$ \\ \hline \hline
\multirow{3}{*}{$14\times 11\times 2$} & MCL \cite{tran2019multilinear}    %
& $84.38$  & $59.08$   & $86.90$    & $84.36$    & $83.93$    \\ \cline{2-7}
& MCLwP (our) %
& $\mathbf{85.84}$  & $\mathbf{59.83}$   & $\mathbf{88.17}$   & $\mathbf{86.99}$    & $\mathbf{87.29}$    \\ \cline{2-7}
	& $\Delta$ (MCLwP$-$MCL) & $\texttt{+}1.46$ & $\texttt{+}0.75$ & $\texttt{+}1.27$ & $\texttt{+}2.63$ & $\texttt{+}3.36$ \\ \hline \hline
$6\times 6\times 1$	& Prior ($\mathsf{P}$)      & $80.71$    & $42.31$     & $75.79$      & $72.24$     & $71.46$     \\ \hline 
$9\times 7\times 1$	& Prior ($\mathsf{P}$)      & $74.77$    & $44.08$     & $81.04$      & $78.70$      & $75.23$  	\\ \hline 
$13\times 12\times 1$	& Prior ($\mathsf{P}$)      & $82.19$  & $53.10$   & $89.75$    & $87.94$   & $87.08$   \\ \hline 
$14\times 11\times 2$	& Prior ($\mathsf{P}$)      & $87.62$  & $61.83$   & $90.83$   & $90.47$    & $91.30$   \\ \hline
\end{tabular}
		}
	\end{center}
\end{table*}

Table \ref{t2} shows the performances of MCL proposed by \cite{tran2019multilinear} and our proposed algorithm MCLwP. The last four rows of Table \ref{t2} present the learning performances of the corresponding prior-generating models. It is clear that with the presence of prior knowledge, our proposed algorithm outperforms MCL for most of the configurations in all five datasets. 

For a simpler problem such as recognizing $10$ objects in CIFAR-10 dataset, the performance gaps between MCLwP and MCL are relatively small. However, when the complexity of the learning problem increases, i.e., when the number of objects or facial identities increases, the differences in performance between MCLwP and MCL are significant. This can be observed by looking at the last row of each measurement configuration, with a direction from left to right (the direction of increasing difficulties of the learning tasks). For example, at configuration $9\times 7\times 1$, moving from CIFAR-10 to CIFAR-100, the improvement changes from $0.67$ to $2.97$, while from CelebA-100 to Celeb-200, then to CelebA-500, the improvement changes from $3.42$ to $4.63$ and to $7.20$, respectively. 

The only setting that MCLwP performs slightly worse than MCL is in CIFAR-10 dataset at the lowest measurement rate ($6\times 6\times 1$). By inspecting the prior-generating models' performances in CIFAR-10 dataset, we can see that the prior-generating model at measurement $6\times 6\times 1$ has very high performance, and even performs far better than its counterpart at a higher measurement rate ($9\times 7\times 1$). Thus, the reason we see the degradation in learning performance might be because there is a huge gap between the teacher's and the student's learning capacity that makes the student model unable to learn effectively. This phenomenon in KD has been observed previously in \cite{mirzadeh2019improved}.

One might make an assumption that lower numbers of measurements always associate with lower learning performances. This, however, is not necessarily true for our prior-generating models having nonlinear CS component with different numbers of downsampling layers. In fact, in our prior-generating models, we use two $2\times 2$ max-pooling layers to reduce the spatial dimensions for configuration $6\times 6\times 1$ while only one for configuration $9\times 7\times 1$. Although better performances for each measurement configuration can be achieved by carefully adjusting the corresponding teacher's capacity as in \cite{mirzadeh2019improved}, it is sufficient for us to use a simple design pattern of autoencoder in order to demonstrate the effectiveness of the proposed MCLwP.

%
%
%

\subsection{Effects of Learning Capacity in Prior-Generating Models}

\begin{table*}[]
	\begin{center}
		\caption{Learning performances of our proposed algorithm trained with two different prior-generating models ($\mathsf{P}$ and $\mathsf{P}^{*}$). $\mathsf{P}^{*}$ denotes the higher-capacity one. The last eight rows show the performances of the prior-generating models for reference.}\label{t3}
		\resizebox{0.85\textwidth}{!}{
			\begin{tabular}{|c|c|c|c|c|c|c|} \hline
				Measurements           & Models & CIFAR-10 & CIFAR-100 & CelebA-100 & CelebA-200 & CelebA-500 \\ \hline \hline
				\multirow{2}{*}{$6\times 6\times 1$} & MCLwP & $\mathbf{57.80}$  & $\mathbf{31.17}$  & $55.31$  & $47.61$  & $\mathbf{42.51}$  \\ \cline{2-7}
				& MCLwP\** & $54.94$  & $30.67$  & $\mathbf{55.56}$  & $\mathbf{48.77}$  & $41.75$ \\ \hline \hline 
				\multirow{2}{*}{$9\times 7\times 1$} & MCLwP & $\mathbf{66.81}$  & $\mathbf{36.60}$  & $75.50$  & $\mathbf{72.02}$  & $\mathbf{68.73}$  \\ \cline{2-7}
				& MCLwP\** & $66.72$  & $36.07$  & $75.77$  & $71.78$  & $68.49$ \\ \hline \hline 
				\multirow{2}{*}{$13\times 12\times 1$} & MCLwP & $77.58$  & $\mathbf{47.92}$  & $87.35$  & $\mathbf{86.39}$  & $85.97$  \\ \cline{2-7}
				& MCLwP\** & $\mathbf{78.27}$  & $47.75$  & $\mathbf{87.50}$  & $85.99$  & $\mathbf{86.10}$ \\ \hline \hline 
				\multirow{2}{*}{$14\times 11\times 2$} & MCLwP & $\mathbf{85.84}$  & $59.83$  & $\mathbf{88.17}$  & $86.99$  & $87.29$  \\ \cline{2-7}
				& MCLwP\** & $85.78$  & $\mathbf{60.83}$  & $87.87$  & $\mathbf{87.09}$  & $\mathbf{87.41}$ \\ \hline \hline 
				\multirow{2}{*}{$6\times 6\times 1$} & Prior $\mathsf{P}$ & $80.71$  & $42.31$  & $75.79$  & $72.24$  & $71.46$  \\ \cline{2-7}
				& Prior $\mathsf{P}^{*}$ & $\mathbf{84.29}$  & $\mathbf{49.47}$  & $\mathbf{79.75}$  & $\mathbf{76.48}$  & $\mathbf{76.74}$  \\ \hline \hline 
				\multirow{2}{*}{$9\times 7\times 1$} & Prior $\mathsf{P}$ & $74.77$  & $44.08$  & $81.04$  & $78.70$  & $75.23$  \\ \cline{2-7}
				& Prior $\mathsf{P}^{*}$ & $\mathbf{82.91}$  & $\mathbf{48.58}$  & $\mathbf{84.25}$  & $\mathbf{82.49}$  & $\mathbf{81.27}$ \\ \hline \hline 
				\multirow{2}{*}{$13\times 12\times 1$} & Prior $\mathsf{P}$ & $82.19$  & $53.10$  & $89.75$  & $87.94$  & $87.08$  \\ \cline{2-7}
				& Prior $\mathsf{P}^{*}$ & $\mathbf{84.53}$  & $\mathbf{57.23}$  & $\mathbf{91.04}$  & $\mathbf{89.22}$  & $\mathbf{89.27}$ \\ \hline \hline 
				\multirow{2}{*}{$14\times 11\times 2$} & Prior $\mathsf{P}$ & $87.62$  & $61.83$  & $90.83$  & $90.47$  & $91.30$  \\ \cline{2-7}
				& Prior $\mathsf{P}^{*}$ & $\mathbf{87.93}$  & $\mathbf{63.69}$  & $\mathbf{91.42}$  & $\mathbf{90.53}$  & $\mathbf{92.11}$ \\ \hline
			\end{tabular}
		}
	\end{center}
\end{table*}

While individual tweaking of each prior-generating model's topology requires elaborate experimentation and is out of the scope of this work, we still conducted a simple set of experiments to study the overall effects when changing the prior-generating models' capacity. In particular, we increased the capacity of the teacher models ($\mathsf{P}$) in Section \ref{S4.3} by adding more convolution layers in the CS and FS components. The set of higher-capacity teachers are denoted as $\mathsf{P}^*$, and the resulting student models are denoted as MCLwP\**. 

Table \ref{t3} shows the learning performances of MCLwP in comparison with MCLwP\**, and $\mathsf{P}$ in comparison with $\mathsf{P}^*$. While there are clear improvements in the learning performance of the teachers when we switch from $\mathsf{P}$ to $\mathsf{P}^*$, we observe mixed behaviors in the corresponding student models. Here we should note that this phenomenon is expected since different measurement configurations in different datasets would require different adjustments (increase or decrease) of the teacher's capacity to ensure the most efficient knowledge distillation. As observed in \cite{mirzadeh2019improved}, the distribution of a student model's performance with respect to different teacher models' capacity has a bell shape. Thus, Table \ref{t3} can act as a guideline whether to increase or decrease the capacity of the prior-generating model in each configuration to maximize the learning performance of its student. 

For example, in CIFAR-10 and CIFAR-100 dataset at measurement $6\times 6\times 1$, increasing the teacher's capacity from $\mathsf{P}$ to $\mathsf{P}^*$ leads to further degradation in the student's performances; thus we should lower the capacity of $\mathsf{P}$ to move toward the bell curve's peak. On the other hand, in CIFAR-100 at $14\times 11\times 2$ or in CelebA-200 at $6\times 6\times 1$, we should further upgrade the capacity of $\mathsf{P}^*$ to possibly obtain better performing student models. As mentioned previously, this type of empirical hill climbing for each measurement configuration requires extensive experiments, however, might be necessary for certain practical applications.

\subsection{Quality of Compressive Domain}
%
%
%
\begin{table*}[t!]
	\begin{center}
		\caption{K-nearest neighbor performances (test accuracy in \%) calculated using compressed measurements produced by MCL \cite{tran2019multilinear} and MCLwP (our proposed algorithm)}\label{t4}
		\resizebox{0.85\textwidth}{!}{
			\begin{tabular}{|c|c|c|c|c|c|c|} \hline
				Measurements           & Models & CIFAR-10 & CIFAR-100 & CelebA-100 & CelebA-200 & CelebA-500 \\ \hline \hline
				\multicolumn{7}{|c|}{\textit{\#neighbors k=5}} \\ \hline \hline
\multirow{2}{*}{$6\times 6\times 1$} & MCL \cite{tran2019multilinear}  & $\mathbf{38.21}$ & $14.60$ & $25.15$ & $19.68$ & $13.75$ \\  \cline{2-7}
& MCLwP (our) & $37.70$ & $\mathbf{15.02}$ & $\mathbf{31.92}$ & $\mathbf{22.81}$ & $\mathbf{15.45}$ \\ \hline \hline
\multirow{2}{*}{$9\times 7\times 1$} & MCL \cite{tran2019multilinear} & $\mathbf{37.50}$ & $\mathbf{13.79}$ & $30.15$ & $25.02$ & $18.37$ \\ \cline{2-7}
& MCLwP (our) & $35.52$ & $12.56$ & $\mathbf{36.77}$ & $\mathbf{29.69}$ & $\mathbf{21.41}$ \\ \hline \hline
				\multicolumn{7}{|c|}{\textit{\#neighbors k=20}} \\ \hline \hline
\multirow{2}{*}{$6\times 6\times 1$} & MCL \cite{tran2019multilinear} & $38.87$ & $14.73$ & $24.21$ & $19.95$ & $14.94$ \\  \cline{2-7}
& MCLwP (our) & $\mathbf{39.36}$ & $\mathbf{15.32}$ & $\mathbf{31.48}$ & $\mathbf{23.46}$ & $\mathbf{17.08}$ \\ \hline \hline
\multirow{2}{*}{$9\times 7\times 1$} & MCL \cite{tran2019multilinear} & $\mathbf{37.16}$ & $13.01$ & $30.21$ & $25.31$ & $19.59$ \\ \cline{2-7}
& MCLwP (our) & $34.93$ & $\mathbf{13.16}$ & $\mathbf{35.60}$ & $\mathbf{29.42}$ & $\mathbf{22.50}$ \\ \hline
			\end{tabular}
		}
	\end{center}
\end{table*}

Although both MCL and MCLwP optimize the compressive learning models in an end-to-end manner, and there is no explicit loss term that regulates the compressive domain, it is still intuitive to expect models with better learning performances to possess better compressed representation at the same measurement rate. In order to quantify the representation produced by the competing algorithms in the compressive domain, we performed K-Nearest-Neighbor classification using the compressed representation at $6\times 6\times 1$ and $9\times 7\times 1$ after training MCL and MCLwP. Table \ref{t4} shows the learning performances with two different neighbor values ($k=5$ and $k=20$). 

It is clear that MCLwP outperforms MCL in the majority of configurations. The performance gaps between MCLwP and MCL are more significant in facial image recognition tasks than in object recognition tasks. Here we should note that Euclidean distance was used to measure the similarity between data points, which might not be the optimal metric which can entirely reflect the semantic similarity and the quality of the compressive domain, especially when our compressive domains possess tensor forms that also encode spatial information. Besides, we cannot compare the performances of K-Nearest-Neighbor on a dataset across different measurements since measurements having larger spatial dimensions potentially retain more spatial variances, and Euclidean distance becomes less effective when measuring the similarity. For example, we can observe a decrease in performance on the CIFAR-10 and CIFAR-100 datasets, although the number of measurements increases from $6\times 6\times 1$ to $9\times 7\times 1$. For this reason, we did not study the performances of K-Nearest-Neighbor at higher measurements. 

\subsection{Effects of Prior Knowledge}\label{S4.6}
%
%
%
\begin{table*}[t!]
	\begin{center}
		\caption{Learning performances with (MCLwP) and without (MCLw/oP) the presence of prior knowledge}\label{t5}
		\resizebox{0.85\textwidth}{!}{
			\begin{tabular}{|c|c|c|c|c|c|c|} \hline
				Measurements           & Models & CIFAR-10 & CIFAR-100 & CelebA-100 & CelebA-200 & CelebA-500 \\ \hline \hline
\multirow{2}{*}{$6\times 6\times 1$} & MCLw/oP & $\mathbf{58.33}$  & $30.27$  & $\mathbf{55.88}$  & $\mathbf{48.49}$  & $42.40$  \\ \cline{2-7}
& MCLwP & $57.80$  & $\mathbf{31.17}$  & $55.31$  & $47.61$  & $\mathbf{42.51}$ \\ \hline \hline
\multirow{2}{*}{$9\times 7\times 1$} & MCLw/oP & $66.55$  & $34.02$  & $73.83$  & $69.02$  & $67.24$  \\ \cline{2-7}
& MCLwP & $\mathbf{66.81}$  & $\mathbf{36.60}$  & $\mathbf{75.50}$  & $\mathbf{72.02}$  & $\mathbf{68.73}$ \\ \hline \hline 
\multirow{2}{*}{$13\times 12\times 1$} & MCLw/oP & $77.49$  & $44.80$  & $86.71$  & $83.12$  & $83.37$  \\ \cline{2-7}
& MCLwP & $\mathbf{77.58}$  & $\mathbf{47.92}$  & $\mathbf{87.35}$  & $\mathbf{86.39}$  & $\mathbf{85.77}$ \\ \hline \hline
\multirow{2}{*}{$14\times 11\times 2$} & MCLw/oP & $84.42$  & $58.09$  & $85.21$  & $84.76$  & $84.00$  \\ \cline{2-7}
& MCLwP & $\mathbf{85.84}$  & $\mathbf{59.83}$  & $\mathbf{88.17}$  & $\mathbf{86.99}$  & $\mathbf{87.09}$ \\ \hline
			\end{tabular}
		}
	\end{center}
\end{table*}

In order to study the effects of exploiting prior knowledge, we conducted two sets of experiments. While MCL and MCLwP models share the same structures of the CS and task-specific neural network component, there are architectural differences in the FS components: in MCL, the FS component performs multilinear transformation while in MCLwP, the synthesis step consists of both convolution and multilinear operations. Thus, in the first set of experiments, we aim to remove the architectural differences between the MCL and MCLwP models in the FS component that can potentially affect the quantification of prior knowledge. This is done by training the architectures which are trained by MCLwP in a similar manner as proposed in \cite{tran2019multilinear}, i.e., without prior knowledge: we first initialized the CS and FS components by minimizing the $l_1$ reconstructed error via stochastic optimization, then trained the whole model with respect to the inference loss. This set of models are denoted as MCLw/oP and the comparisons between MCLwP and MCLw/oP are presented in Table \ref{t5}. Although we observe mixed behaviors at measurement $6\times 6\times 1$, it is obvious that the presence of prior knowledge leads to improved learning performances without any architectural differences. 

In the second set of experiments, we studied the effect of different knowledge transfer stages in MCLwP by either skipping or performing a particular knowledge transfer stage. This setup leads to $2^3=8$ different variants of the training procedure, which are presented in Table \ref{t6}. Here we should note that when skipping the last transfer stage, we only discarded the distillation loss term and still trained the models with respect to the inference loss, instead of discarding the entire 3\textsuperscript{rd} stage. 

The first row of each dataset shows the results when all knowledge transfer activities are skipped. In other words, the architectures are initialized with a standard neural network initialization technique \cite{glorot2010understanding} and trained only with the inference loss. It is clear that this training procedure produces the worst performing models.

The next three rows in each dataset represent the cases where we performed only one of the knowledge transfer stages. In this scenario, conducting only the 2\textsuperscript{nd} stage has noticeably better results than the other two cases, indicating the importance of the 2\textsuperscript{nd} knowledge transfer stage. 

Regarding the cases when one of the stages is skipped, we can also observe a homogeneous phenomenon across different measurements and different datasets that skipping the 1\textsuperscript{st} stage leads to the least degradation. In fact, with this training procedure, we can obtain performances relatively close to the original MCLwP. During the 2\textsuperscript{nd} knowledge transfer stage, the FS component ($\mathsf{D}$) is updated in conjunction with the CS component ($\mathsf{E}$) to mimic the features $\mathsf{D_P}(\mathsf{E_P}(\mathcal{Y}))$ synthesized by the prior-generating model, which might imply an implicit knowledge transfer from $\mathsf{E_P}$ to $\mathsf{E}$, thus might explain why skipping the 1\textsuperscript{st} transfer stage only leads to minor degradations in performance. Overall, performing all three knowledge transfer stages yields the best results.
%
%
%

\begin{table*}[t!]
	\begin{center}
		\caption{Learning performances (test accuracy in \%) of our proposed algorithm (MCLwP) when different knowledge transfer stages are skipped. The checkmark indicates when a particular transfer stage was conducted}\label{t6}
		\resizebox{0.77\textwidth}{!}{
			\begin{tabular}{|c|c|c|c|c|c|c|} \hline
\makecell{Transfer \\ $\mathsf{E_P}$} & \makecell{Transfer \\ $\mathsf{D_P}(\mathsf{E_P})$} & \makecell{Transfer \\ $\mathsf{N_P}(\mathsf{D_P}(\mathsf{E_P}))$} & $6\times 6\times 1$ & $9\times 7\times 1$ & $13\times 12\times 1$ & $14\times 11\times 2$ \\ \hline \hline
\multicolumn{7}{|c|}{CIFAR-10} \\ \hline \hline
&  &  & $52.74$ & $56.16$ & $57.36$ & $66.02$ \\ \hline
\checkmark &  &  & $53.99$ & $57.68$ & $61.63$ & $66.82$ \\ \hline
& \checkmark &  & $57.91$ & $65.97$ & $77.12$ & $84.86$ \\ \hline
&  & \checkmark & $54.38$ & $59.31$ & $61.23$ & $67.74$ \\ \hline
\checkmark & \checkmark &  & $57.17$ & $65.97$ & $77.16$ & $84.80$ \\ \hline
& \checkmark & \checkmark & $57.75$ & $66.70$ & $\mathbf{77.89}$ & $85.10$ \\ \hline
\checkmark &  & \checkmark & $54.77$ & $59.04$ & $63.07$ & $68.40$ \\ \hline
\checkmark & \checkmark & \checkmark & $\mathbf{57.80}$ & $\mathbf{66.81}$ & $77.58$ & $\mathbf{85.84}$ \\ \hline \hline
\multicolumn{7}{|c|}{CIFAR-100} \\ \hline \hline
 &  &  & $25.13$  & $27.00$  & $28.81$  & $37.77$  \\ \hline
\checkmark &  &  & $26.39$  & $28.78$  & $30.03$  & $40.84$  \\ \hline
& \checkmark &  & $29.52$  & $35.31$  & $46.36$  & $58.96$  \\ \hline
&  & \checkmark & $27.81$  & $31.16$  & $30.76$  & $41.04$  \\ \hline
\checkmark & \checkmark &  & $29.67$  & $35.45$  & $46.42$  & $58.90$  \\ \hline
& \checkmark & \checkmark & $31.12$  & $36.41$  & $47.73$  & $\mathbf{59.89}$  \\ \hline
\checkmark &  & \checkmark & $28.14$  & $31.84$  & $31.87$  & $43.43$  \\ \hline
\checkmark & \checkmark & \checkmark & $\mathbf{31.17}$  & $\mathbf{36.60}$  & $\mathbf{47.92}$  & $59.83$ \\ \hline \hline
\multicolumn{7}{|c|}{CelebA-100} \\ \hline \hline
 &  &  & $33.58$ & $47.19$ & $55.75$  & $57.13$  \\ \hline
\checkmark &  &  & $47.75$  & $59.04$  & $65.08$  & $65.10$  \\ \hline
& \checkmark &  & $53.88$  & $74.77$  & $86.35$  & $87.21$  \\ \hline
&  & \checkmark & $46.42$  & $48.88$  & $52.71$  & $58.77$  \\ \hline
\checkmark & \checkmark &  & $54.40$  & $75.17$  & $86.96$  & $87.94$  \\ \hline
& \checkmark & \checkmark & $53.96$  & $74.06$  & $86.60$  & $87.50$  \\ \hline
\checkmark &  & \checkmark & $48.88$  & $33.19$  & $65.67$  & $65.77$  \\ \hline
\checkmark & \checkmark & \checkmark & $\mathbf{55.31}$  & $\mathbf{75.50}$  & $\mathbf{87.35}$  & $\mathbf{88.17}$ \\ \hline \hline
\multicolumn{7}{|c|}{CelebA-200} \\ \hline \hline
 &  &  & $38.81$  & $36.21$  & $46.73$  & $53.96$  \\ \hline 
\checkmark &  &  & $42.23$  & $56.17$  & $51.24$  & $61.00$  \\ \hline 
& \checkmark &  & $46.68$  & $70.69$  & $85.31$  & $86.41$  \\ \hline 
&  & \checkmark & $42.61$  & $49.27$  & $54.28$  & $56.62$  \\ \hline 
\checkmark & \checkmark &  & $46.73$  & $71.50$  & $86.06$  & $86.40$  \\ \hline 
& \checkmark & \checkmark & $47.53$  & $71.52$  & $85.86$  & $86.42$  \\ \hline 
\checkmark &  & \checkmark & $43.59$  & $58.72$  & $61.88$  & $63.37$  \\ \hline 
\checkmark & \checkmark & \checkmark & $\mathbf{47.61}$  & $\mathbf{72.02}$  & $\mathbf{86.39}$  & $\mathbf{86.99}$ \\ \hline \hline 
\multicolumn{7}{|c|}{CelebA-500} \\ \hline \hline
 &  &  & $33.60$  & $46.46$  & $54.40$  & $53.03$  \\ \hline
\checkmark &  &  & $37.46$  & $51.31$  & $61.53$  & $61.48$  \\ \hline
& \checkmark &  & $40.68$  & $67.70$  & $85.31$  & $86.46$  \\ \hline
&  & \checkmark & $37.01$  & $53.06$  & $57.45$  & $55.92$  \\ \hline
\checkmark & \checkmark &  & $41.61$  & $67.81$  & $85.47$  & $86.60$  \\ \hline
& \checkmark & \checkmark & $41.89$  & $\mathbf{68.78}$  & $85.95$  & $87.27$  \\ \hline
\checkmark &  & \checkmark & $39.11$  & $56.91$  & $64.83$  & $63.22$  \\ \hline
\checkmark & \checkmark & \checkmark & $\mathbf{42.51}$  & $68.73$  & $\mathbf{85.97}$  & $\mathbf{87.29}$ \\ \hline
			\end{tabular}
		}
	\end{center}
\end{table*}

\subsection{Semi-supervised Learning}\label{S4.7}
%
%
%
%
%
\begin{table*}[t!]
	\begin{center}
		\caption{Learning performances in Semi-supervised Setting of MCL \cite{tran2019multilinear}, MCLwP (trained with labeled data only), and MCLwP-S (trained with labeled and unlabeled data). The last eight rows show the performances of the corresponding prior-generating models ($\mathsf{P}$ and $\mathsf{P}\textnormal{-}\mathsf{S}$ which provide prior knowledge for MCLwP and MCLwP-S, respectively.)}\label{t7}
		\resizebox{0.7\textwidth}{!}{
			\begin{tabular}{|c|c|c|c|c|} \hline
				Measurements           & Models & CIFAR-10S & CIFAR-100S & CelebA-500S  \\ \hline \hline
\multirow{3}{*}{$6\times 6\times 1$} & MCL & $\mathbf{51.19}$  & $17.59$  & $12.93$  \\ \cline{2-5}
& MCLwP & $47.78$  & $18.35$  & $17.70$  \\ \cline{2-5}
& MCLwP-S & $50.13$  & $\mathbf{19.63}$  & $\mathbf{21.91}$ \\ \hline \hline
\multirow{3}{*}{$9\times 7\times 1$} & MCL & $56.55$  & $22.86$  & $27.46$  \\ \cline{2-5}
& MCLwP & $55.75$  & $20.05$  & $25.72$  \\ \cline{2-5}
& MCLwP-S & $\mathbf{57.59}$  & $\mathbf{23.03}$  & $\mathbf{34.16}$ \\ \hline \hline
\multirow{3}{*}{$13\times 12\times 1$} & MCL & $67.23$  & $31.12$  & $49.09$  \\ \cline{2-5}
& MCLwP & $67.13$  & $28.97$  & $44.56$  \\ \cline{2-5}
& MCLwP-S & $\mathbf{69.37}$  & $\mathbf{32.67}$  & $\mathbf{51.55}$ \\ \hline \hline
\multirow{3}{*}{$14\times 11\times 2$} & MCL & $75.31$  & $41.10$  & $41.47$  \\ \cline{2-5}
& MCLwP & $75.86$  & $41.55$  & $42.07$  \\ \cline{2-5}
& MCLwP-S & $\mathbf{78.03}$  & $\mathbf{43.87}$  & $\mathbf{58.34}$ \\ \hline \hline
\multirow{2}{*}{$6\times 6\times 1$} & Prior $\mathsf{P}$ & $\mathbf{66.81}$  & $25.88$  & $27.32$  \\ \cline{2-5}
& Prior $\mathsf{P}\textnormal{-}\mathsf{S}$ & $63.96$  & $\mathbf{27.00}$  & $\mathbf{32.79}$  \\ \hline \hline
\multirow{2}{*}{$9\times 7\times 1$} & Prior $\mathsf{P}$ & $62.29$  & $27.46$  & $26.68$  \\ \cline{2-5}
& Prior $\mathsf{P}\textnormal{-}\mathsf{S}$ & $\mathbf{64.44}$  & $\mathbf{29.11}$  & $\mathbf{38.76}$ \\ \hline \hline
\multirow{2}{*}{$13\times 12\times 1$} & Prior $\mathsf{P}$ & $70.95$  & $33.97$  & $47.69$  \\ \cline{2-5}
& Prior $\mathsf{P}\textnormal{-}\mathsf{S}$ & $\mathbf{73.28}$  & $\mathbf{37.11}$  & $\mathbf{54.37}$ \\ \hline \hline
\multirow{2}{*}{$14\times 11\times 2$} & Prior $\mathsf{P}$ & $76.99$  & $43.05$  & $45.97$  \\ \cline{2-5}
& Prior $\mathsf{P}\textnormal{-}\mathsf{S}$ & $\mathbf{79.70}$  & $\mathbf{46.85}$  & $\mathbf{63.83}$ \\ \hline 
			\end{tabular}
		}
	\end{center}
\end{table*}

Finally, we present the empirical results in the semi-supervised setting. Here we should note that although the total amount of data is large, only a small fraction has labels for training. The competing algorithms include MCL \cite{tran2019multilinear} and MCLwP, both of which were trained with the labeled data only, and MCLwP-S, the semi-supervised extension of MCLwP which takes advantages of the unlabeled data in addition to the labeled set. We denote $\mathsf{P}$ and $\mathsf{P}\textnormal{-}\mathsf{S}$ the corresponding prior-generating models of MCLwP and MCLwP-S, respectively. The results are shown in Table \ref{t7}. 

It is clear that without sufficient data, the prior-generating models ($\mathsf{P}$) cannot effectively transfer the knowledge to their student models (MCLwP) as can be seen by the inferior performances of MCLwP compared to MCL in half of the cases. However, when unlabeled data is utilized during the training process as proposed in our MCLwP-S algorithm, it is obvious that not only the prior-generating models ($\mathsf{P}\textnormal{-}\mathsf{S}$) improve in performance but also their student models (MCLwP-S). The enhancement of the knowledge transfer process (via additional data) and prior-generating models (via self-labeling procedure) results in MCLwP-S having the best performances among the competing algorithms.

\section{Conclusions}\label{S5}
In this work, we proposed a novel methodology to find and incorporate data-dependent prior knowledge into the training process of MCL models. In addition to the traditional supervised learning setting, we also proposed a semi-supervised adaptation that enables our methodology to take advantage of unlabeled data that comes from similar distributions of the signal of interest. Although we limited our investigation to MCL framework, the proposals presented in this work are sufficiently generic to be applicable to any CL systems. With extensive sets of experiments, we demonstrated the effectiveness of our algorithms to train MCL models in comparison with the previously proposed algorithm and provided insights into different aspects of the proposed methodologies. 

\section{Acknowledgement}
This project has received funding from the European Union's Horizon 2020 research and innovation programme under grant agreement No 871449 (OpenDR). This publication reflects the authors’ views only. The European Commission is not responsible for any use that may be made of the information it contains.

\bibliography{reference}

\begin{thebibliography}{10}

\bibitem{candes2008introduction}
E.~J. Cand{\`e}s and M.~B. Wakin, ``An introduction to compressive sampling [a
  sensing/sampling paradigm that goes against the common knowledge in data
  acquisition],'' {\em IEEE signal processing magazine}, vol.~25, no.~2,
  pp.~21--30, 2008.

\bibitem{caiafa2013multidimensional}
C.~F. Caiafa and A.~Cichocki, ``Multidimensional compressed sensing and their
  applications,'' {\em Wiley Interdisciplinary Reviews: Data Mining and
  Knowledge Discovery}, vol.~3, no.~6, pp.~355--380, 2013.

\bibitem{august2013compressive}
Y.~August, C.~Vachman, Y.~Rivenson, and A.~Stern, ``Compressive hyperspectral
  imaging by random separable projections in both the spatial and the spectral
  domains,'' {\em Applied optics}, vol.~52, no.~10, pp.~D46--D54, 2013.

\bibitem{patel2010compressed}
V.~M. Patel, G.~R. Easley, D.~M. Healy~Jr, and R.~Chellappa, ``Compressed
  synthetic aperture radar,'' {\em IEEE Journal of selected topics in signal
  processing}, vol.~4, no.~2, pp.~244--254, 2010.

\bibitem{irarrazabal1995fast}
P.~Irarrazabal and D.~G. Nishimura, ``Fast three dimensional magnetic resonance
  imaging,'' {\em Magnetic resonance in medicine}, vol.~33, no.~5,
  pp.~656--662, 1995.

\bibitem{candes2006stable}
E.~J. Candes, J.~K. Romberg, and T.~Tao, ``Stable signal recovery from
  incomplete and inaccurate measurements,'' {\em Communications on Pure and
  Applied Mathematics: A Journal Issued by the Courant Institute of
  Mathematical Sciences}, vol.~59, no.~8, pp.~1207--1223, 2006.

\bibitem{donoho2006compressed}
D.~L. Donoho {\em et~al.}, ``Compressed sensing,'' {\em IEEE Transactions on
  information theory}, vol.~52, no.~4, pp.~1289--1306, 2006.

\bibitem{mota2017compressed}
J.~F. Mota, N.~Deligiannis, and M.~R. Rodrigues, ``Compressed sensing with
  prior information: Strategies, geometry, and bounds,'' {\em IEEE Transactions
  on Information Theory}, vol.~63, no.~7, pp.~4472--4496, 2017.

\bibitem{stankovic2009compressive}
V.~Stankovi{\'c}, L.~Stankovi{\'c}, and S.~Cheng, ``Compressive image sampling
  with side information,'' in {\em 2009 16th IEEE International Conference on
  Image Processing (ICIP)}, pp.~3037--3040, IEEE, 2009.

\bibitem{kang2009distributed}
L.-W. Kang and C.-S. Lu, ``Distributed compressive video sensing,'' in {\em
  2009 IEEE International Conference on Acoustics, Speech and Signal
  Processing}, pp.~1169--1172, IEEE, 2009.

\bibitem{charles2011sparsity}
A.~Charles, M.~S. Asif, J.~Romberg, and C.~Rozell, ``Sparsity penalties in
  dynamical system estimation,'' in {\em 2011 45th annual conference on
  information sciences and systems}, pp.~1--6, IEEE, 2011.

\bibitem{baron2005distributed}
D.~Baron, M.~B. Wakin, M.~F. Duarte, S.~Sarvotham, and R.~G. Baraniuk,
  ``Distributed compressed sensing,'' 2005.

\bibitem{cevher2008compressive}
V.~Cevher, A.~Sankaranarayanan, M.~F. Duarte, D.~Reddy, R.~G. Baraniuk, and
  R.~Chellappa, ``Compressive sensing for background subtraction,'' in {\em
  European Conference on Computer Vision}, pp.~155--168, Springer, 2008.

\bibitem{trocan2010disparity}
M.~Trocan, T.~Maugey, J.~E. Fowler, and B.~Pesquet-Popescu,
  ``Disparity-compensated compressed-sensing reconstruction for multiview
  images,'' in {\em 2010 IEEE International Conference on Multimedia and Expo},
  pp.~1225--1229, IEEE, 2010.

\bibitem{vaswani2010modified}
N.~Vaswani and W.~Lu, ``Modified-cs: Modifying compressive sensing for problems
  with partially known support,'' {\em IEEE Transactions on Signal Processing},
  vol.~58, no.~9, pp.~4595--4607, 2010.

\bibitem{renna2014reconstruction}
F.~Renna, R.~Calderbank, L.~Carin, and M.~R. Rodrigues, ``Reconstruction of
  signals drawn from a gaussian mixture via noisy compressive measurements,''
  {\em IEEE Transactions on Signal Processing}, vol.~62, no.~9, pp.~2265--2277,
  2014.

\bibitem{khajehnejad2011analyzing}
M.~A. {Khajehnejad}, W.~{Xu}, A.~S. {Avestimehr}, and B.~{Hassibi}, ``Analyzing
  weighted $\ell_1$minimization for sparse recovery with nonuniform sparse
  models,'' {\em IEEE Transactions on Signal Processing}, vol.~59,
  pp.~1985--2001, May 2011.

\bibitem{stojnic2009reconstruction}
M.~Stojnic, F.~Parvaresh, and B.~Hassibi, ``On the reconstruction of
  block-sparse signals with an optimal number of measurements,'' {\em IEEE
  Transactions on Signal Processing}, vol.~57, no.~8, pp.~3075--3085, 2009.

\bibitem{eldar2009robust}
Y.~C. Eldar and M.~Mishali, ``Robust recovery of signals from a structured
  union of subspaces,'' {\em IEEE Transactions on Information Theory}, vol.~55,
  no.~11, pp.~5302--5316, 2009.

\bibitem{calderbank2012finding}
R.~Calderbank and S.~Jafarpour, ``Finding needles in compressed haystacks,'' in
  {\em 2012 IEEE International Conference on Acoustics, Speech and Signal
  Processing (ICASSP)}, pp.~3441--3444, IEEE, 2012.

\bibitem{davenport2007smashed}
M.~A. Davenport, M.~F. Duarte, M.~B. Wakin, J.~N. Laska, D.~Takhar, K.~F.
  Kelly, and R.~G. Baraniuk, ``The smashed filter for compressive
  classification and target recognition,'' in {\em Computational Imaging V},
  vol.~6498, p.~64980H, International Society for Optics and Photonics, 2007.

\bibitem{davenport2010signal}
M.~A. Davenport, P.~Boufounos, M.~B. Wakin, R.~G. Baraniuk, {\em et~al.},
  ``Signal processing with compressive measurements.,'' {\em J. Sel. Topics
  Signal Processing}, vol.~4, no.~2, pp.~445--460, 2010.

\bibitem{reboredo2013compressive}
H.~Reboredo, F.~Renna, R.~Calderbank, and M.~R. Rodrigues, ``Compressive
  classification,'' in {\em 2013 IEEE International Symposium on Information
  Theory}, pp.~674--678, IEEE, 2013.

\bibitem{lohit2015reconstruction}
S.~Lohit, K.~Kulkarni, P.~Turaga, J.~Wang, and A.~C. Sankaranarayanan,
  ``Reconstruction-free inference on compressive measurements,'' in {\em
  Proceedings of the IEEE Conference on Computer Vision and Pattern Recognition
  Workshops}, pp.~16--24, 2015.

\bibitem{adler2016compressed}
A.~Adler, M.~Elad, and M.~Zibulevsky, ``Compressed learning: A deep neural
  network approach,'' {\em arXiv preprint arXiv:1610.09615}, 2016.

\bibitem{tran2019multilinear}
D.~T. Tran, M.~Yamac, A.~Degerli, M.~Gabbouj, and A.~Iosifidis, ``Multilinear
  compressive learning,'' {\em arXiv preprint arXiv:1905.07481}, 2019.

\bibitem{yamac2019reversible}
M.~Yamac, M.~Ahishali, N.~Passalis, J.~Raitoharju, B.~Sankur, and M.~Gabbouj,
  ``Reversible privacy preservation using multi-level encryption and
  compressive sensing,'' {\em arXiv preprint arXiv:1906.08713}, 2019.

\bibitem{mohassel2017secureml}
P.~Mohassel and Y.~Zhang, ``Secureml: A system for scalable privacy-preserving
  machine learning,'' in {\em 2017 IEEE Symposium on Security and Privacy
  (SP)}, pp.~19--38, IEEE, 2017.

\bibitem{reboredo2013projections}
H.~Reboredo, F.~Renna, R.~Calderbank, and M.~R. Rodrigues, ``Projections
  designs for compressive classification,'' in {\em 2013 IEEE Global Conference
  on Signal and Information Processing}, pp.~1029--1032, IEEE, 2013.

\bibitem{baheti2008adaptive}
P.~K. Baheti and M.~A. Neifeld, ``Adaptive feature-specific imaging: a face
  recognition example,'' {\em Applied optics}, vol.~47, no.~10, pp.~B21--B31,
  2008.

\bibitem{lohit2016direct}
S.~Lohit, K.~Kulkarni, and P.~Turaga, ``Direct inference on compressive
  measurements using convolutional neural networks,'' in {\em 2016 IEEE
  International Conference on Image Processing (ICIP)}, pp.~1913--1917, IEEE,
  2016.

\bibitem{hollis2018compressed}
B.~Hollis, S.~Patterson, and J.~Trinkle, ``Compressed learning for tactile
  object recognition,'' {\em IEEE Robotics and Automation Letters}, vol.~3,
  no.~3, pp.~1616--1623, 2018.

\bibitem{deugerli2018compressively}
A.~De{\u{g}}erli, S.~Aslan, M.~Yamac, B.~Sankur, and M.~Gabbouj,
  ``Compressively sensed image recognition,'' in {\em 2018 7th European
  Workshop on Visual Information Processing (EUVIP)}, pp.~1--6, IEEE, 2018.

\bibitem{zisselman2018compressed}
E.~Zisselman, A.~Adler, and M.~Elad, ``Compressed learning for image
  classification: A deep neural network approach,'' {\em Processing, Analyzing
  and Learning of Images, Shapes, and Forms}, vol.~19, p.~1, 2018.

\bibitem{bengio2013representation}
Y.~Bengio, A.~Courville, and P.~Vincent, ``Representation learning: A review
  and new perspectives,'' {\em IEEE transactions on pattern analysis and
  machine intelligence}, vol.~35, no.~8, pp.~1798--1828, 2013.

\bibitem{hinton2015distilling}
G.~Hinton, O.~Vinyals, and J.~Dean, ``Distilling the knowledge in a neural
  network,'' {\em arXiv preprint arXiv:1503.02531}, 2015.

\bibitem{baraniuk2009random}
R.~G. Baraniuk and M.~B. Wakin, ``Random projections of smooth manifolds,''
  {\em Foundations of computational mathematics}, vol.~9, no.~1, pp.~51--77,
  2009.

\bibitem{he2016identity}
K.~He, X.~Zhang, S.~Ren, and J.~Sun, ``Identity mappings in deep residual
  networks,'' in {\em European conference on computer vision}, pp.~630--645,
  Springer, 2016.

\bibitem{zhang2018deep}
Y.~Zhang, T.~Xiang, T.~M. Hospedales, and H.~Lu, ``Deep mutual learning,'' in
  {\em Proceedings of the IEEE Conference on Computer Vision and Pattern
  Recognition}, pp.~4320--4328, 2018.

\bibitem{romero2014fitnets}
A.~Romero, N.~Ballas, S.~E. Kahou, A.~Chassang, C.~Gatta, and Y.~Bengio,
  ``Fitnets: Hints for thin deep nets,'' {\em arXiv preprint arXiv:1412.6550},
  2014.

\bibitem{yim2017gift}
J.~Yim, D.~Joo, J.~Bae, and J.~Kim, ``A gift from knowledge distillation: Fast
  optimization, network minimization and transfer learning,'' in {\em
  Proceedings of the IEEE Conference on Computer Vision and Pattern
  Recognition}, pp.~4133--4141, 2017.

\bibitem{yu2017visual}
R.~Yu, A.~Li, V.~I. Morariu, and L.~S. Davis, ``Visual relationship detection
  with internal and external linguistic knowledge distillation,'' in {\em
  Proceedings of the IEEE International Conference on Computer Vision},
  pp.~1974--1982, 2017.

\bibitem{anil2018large}
R.~Anil, G.~Pereyra, A.~Passos, R.~Ormandi, G.~E. Dahl, and G.~E. Hinton,
  ``Large scale distributed neural network training through online
  distillation,'' {\em arXiv preprint arXiv:1804.03235}, 2018.

\bibitem{triguero2015self}
I.~Triguero, S.~Garc{\'\i}a, and F.~Herrera, ``Self-labeled techniques for
  semi-supervised learning: taxonomy, software and empirical study,'' {\em
  Knowledge and Information systems}, vol.~42, no.~2, pp.~245--284, 2015.

\bibitem{krizhevsky2009learning}
A.~Krizhevsky and G.~Hinton, ``Learning multiple layers of features from tiny
  images,'' tech. rep., Citeseer, 2009.

\bibitem{liu2015faceattributes}
Z.~Liu, P.~Luo, X.~Wang, and X.~Tang, ``Deep learning face attributes in the
  wild,'' in {\em Proceedings of International Conference on Computer Vision
  (ICCV)}, 2015.

\bibitem{kingma2014adam}
D.~P. Kingma and J.~Ba, ``Adam: A method for stochastic optimization,'' {\em
  arXiv preprint arXiv:1412.6980}, 2014.

\bibitem{mirzadeh2019improved}
S.-I. Mirzadeh, M.~Farajtabar, A.~Li, and H.~Ghasemzadeh, ``Improved knowledge
  distillation via teacher assistant: Bridging the gap between student and
  teacher,'' {\em arXiv preprint arXiv:1902.03393}, 2019.

\bibitem{glorot2010understanding}
X.~Glorot and Y.~Bengio, ``Understanding the difficulty of training deep
  feedforward neural networks,'' in {\em Proceedings of the thirteenth
  international conference on artificial intelligence and statistics},
  pp.~249--256, 2010.

\end{thebibliography}
\bibliographystyle{ieeetr}

\end{document}